\documentclass[letterpaper]{article} %
\usepackage[preprint]{aaai2027}  %
\usepackage[hyphens]{url}  %
\usepackage{graphicx} %
\usepackage{natbib}  %
\usepackage{caption} %
\usepackage{algorithm}
\usepackage{algorithmic}

\usepackage{newfloat}
\usepackage{listings}
\DeclareCaptionStyle{ruled}{labelfont=normalfont,labelsep=colon,strut=off} %
\floatstyle{ruled}
\newfloat{listing}{tb}{lst}{}
\floatname{listing}{Listing}

\usepackage{booktabs}

\usepackage{mathtools} %
\usepackage{nicefrac}       %
\usepackage{microtype}      %
\usepackage[dvipsnames]{xcolor}         %
\usepackage{amssymb}
\usepackage{amsmath}
\usepackage{subcaption}
\usepackage{enumitem}
\usepackage{amsthm}
\usepackage{multirow}
\usepackage[most]{tcolorbox}

\newif\ifshowComments
\showCommentstrue
\showCommentsfalse

\ifshowComments
\newcommand{\al}[1]{{\color{red}{#1}}}
\newcommand{\ee}[1]{{\color{blue}{#1}}}
\else
\newcommand{\al}[1]{}
\newcommand{\ee}[1]{}
\fi

\newcommand{\bx}{\mathbf{x}}
\newcommand{\bM}{\mathbf{M}}
\newcommand{\bz}{\mathbf{z}}
\newcommand{\bD}{\mathbf{D}}
\newcommand{\bR}{\mathbf{R}}
\newcommand{\R}{\mathbb{R}}

\newtheorem{hypothesis}{Hypothesis}
\newtcolorbox{hypothesisbox}{
  colback=yellow!10,
  colframe=YellowOrange!60,
  boxrule=0.5pt,
  arc=2pt,
  left=6pt,
  right=6pt,
  top=6pt,
  bottom=6pt
}

\newtcolorbox{setupbox}{
  colback=blue!7,
  colframe=blue!60,
  boxrule=0.5pt,
  arc=3pt,
  left=6pt,
  right=6pt,
  top=6pt,
  bottom=6pt
}

\newtheorem*{setupinner}{Setup}

\definecolor{EquivariantGreen}{HTML}{82B366}
\definecolor{InvariantPurple}{HTML}{A680B8}
\definecolor{NNBlue}{HTML}{0066CC}

\definecolor{CameraReadyUpdate}{HTML}{000000}

\title{Equivariant Sparse Autoencoders: Mechanistic Interpretability of Neural Networks on Symmetric Data}
\author{
  Ege Erdogan\corresponding,
  Ana Lucic
}
\affiliations{
  University of Amsterdam\\
  e.erdogan@uva.nl
}

\begin{document}

\maketitle

\begin{abstract}

  Machine learning (ML) models achieve remarkable performance but remain hard to interpret due to their scale and complexity. In particular, their activations entangle many concepts into fewer dimensions, a phenomenon known as superposition. Mechanistic interpretability methods such as \textit{sparse autoencoders} (SAEs) can disentangle these dense activations into sparse sums of interpretable \textit{features}, but SAEs suffer from unidentifiability: different explanations can fit the data equally well without necessarily being more interpretable or faithful to the underlying model.
  We show that this problem is exacerbated by data symmetries such as rotations that are prevalent in scientific domains. We extend the Linear Representation Hypothesis, the theory behind SAEs, to account for symmetries and show on synthetic as well as real-world scientific datasets and models that the resulting \textit{Equivariant SAEs} can (1) avoid the pitfalls of existing SAEs on symmetric data and (2) discover features more useful for downstream tasks despite worse reconstructions. Our results show that incorporating the correct priors in SAEs can significantly improve their usefulness while highlighting that reconstruction quality can be inversely correlated with feature usefulness under symmetries, cautioning against its use as a key measure of interpretability.

  \includegraphics[height=1em]{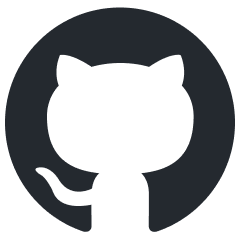} \url{https://github.com/ege-erdogan/equivariant-sae}
\end{abstract}

\section{Introduction}

\begin{figure*}[t!]
  \centering
  \includegraphics[width=0.9\linewidth]{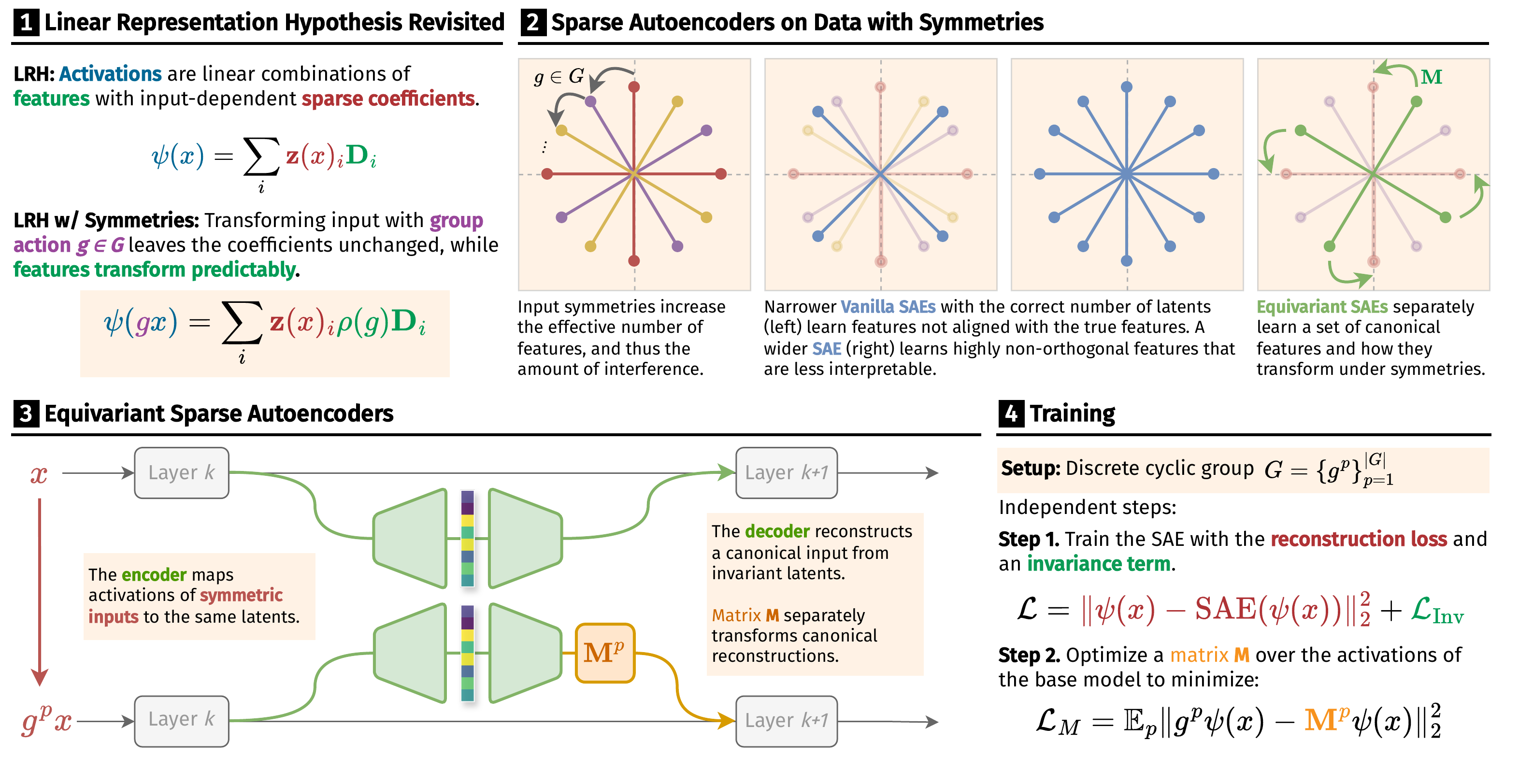}
  \caption{\label{fig:method}
    \textbf{Overview.} (1) We adapt the Linear Representation Hypothesis to input symmetries. (2) This highlights the shortcomings of vanilla SAEs on data with symmetries. (3) We address those by designing Equivariant SAEs: an invariant SAE and a separate linear transformation that models how input symmetries are reflected in activations, (4) trained in two independent steps.
  }
\end{figure*}

Machine learning (ML) models are increasingly used for scientific problems, from higher-level data analysis and hypothesis generation \cite{gottweisAICoscientist2025} to lower-level emulation of physical processes such as protein folding \cite{abramsonAccurateStructurePrediction2024}. However, they are often uninterpretable black boxes, and being able to interpret their internals would not only ensure they are scientifically reliable and controllable, but also potentially lead to novel insights about the underlying domains they simulate \citep{luMechanismsAIProtein2026}.

The field of \textit{mechanistic interpretability} (MI) targets this problem (see survey by \citet{bereskaMechanisticInterpretabilityAI2024}), aiming to explain mechanisms inside ML models using primitives such as \textit{features}.
A widely-used MI tool, adapted from the sparse coding literature \citep{olshausenSparseCodingOvercomplete1997}, is the \textit{sparse autoencoder} (SAE) \citep{cunninghamSparseAutoencodersFind2023,bricken2023monosemanticity} which decomposes dense neural network activations into sparse sums of more interpretable features.

Progress in SAEs has largely been driven by applications on language models \citep{mcdougallGemmaScope22025}. While they have been applied to scientific models in domains such as proteins \citep{parsanInterpretableStructurePrediction2025,simonInterPLMDiscoveringInterpretable2025}, cell images \cite{dasdelenCytoSAEInterpretableCell2025, donhauserScientificDiscoveryDictionary2025}, and molecules \citep{varadiCircuitsFeaturesHeuristics2025} as well, aspects such as symmetries in the data are overlooked.
Developing symmetry-aware models has become an active research area \citep{bronsteinGeometricDeepLearning2021} that can lead to more data-efficient models \citep{brehmerDoesEquivarianceMatter2024}, but so far there is limited work on SAEs in this context.

In this paper, our main research question is: \textit{How should mechanistic interpretability methods such as SAEs account for symmetries, and what benefits does doing so provide?}
Our contributions are:
\begin{itemize}[leftmargin=*]
  \item We extend the Linear Representation Hypothesis \citep{elhage2022superposition} that motivates SAEs, to account for symmetries.
  \item We design Equivariant SAEs that respect these symmetries and avoid various pitfalls of existing SAEs.
  \item We show that our Equivariant SAEs (1) achieve better feature recovery/detection on our toy model and (2) learn features more useful for downstream probing tasks across 3 model architectures on both synthetic and real data.
  \item We provide empirical evidence across all our settings that reconstruction quality can be inversely correlated with feature usefulness, a finding that contrasts the common paradigm of using reconstruction quality as the primary metric.
\end{itemize}

Together, our results highlight the potential benefit of aligning the priors of interpretability methods with our knowledge of the data they operate on, and the risk of primarily relying on reconstruction metrics to evaluate them.

\section{Background} \label{sec:background}

\textbf{Symmetries \& Groups.}
Scientific data often involve \textit{symmetries}: transformations such as rotations that preserve certain properties while transforming others accordingly.
These symmetries are formalized through \textit{groups}. A group $G$ is a set with an associative binary operator, an identity element, and inverses for each element. Groups can be \textit{discrete} (e.g., cyclic group $C_n$) or \textit{continuous} (e.g., rotations in $n$-dimensional space).
Groups \textit{act} on sets $X$ via $x \mapsto gx$ with $g \in G$. The \textit{orbit} of $x \in X$ is $\{ gx : g \in G\}$. A function $f: X \to Y$ is \textit{invariant} if $f(gx) = f(x)$ or \textit{equivariant} if $f(gx) = gf(x)$ for all $g \in G$.

\textbf{Linear Representation Hypothesis and SAEs.}
Neural networks encode more concepts than they have neurons for \cite{elhage2022superposition}. The Linear Representation Hypothesis (LRH) formalized by \cite{elhage2022superposition} states that features are directions in activation space and activations are sparse sums of features: for input $\bx \in \R^n$ and activations $\psi(\bx) \in \R^m$,
$\psi(\bx) = \sum_{i=1}^{C} \bz(\bx)_i \bD_i$
where $\bz: \R^n \to \R^C_{\geq 0}$ gives feature magnitudes and $\bD_i$'s are approximately orthonormal feature vectors, with $\vert \text{supp}(\bz(x)) \vert \ll C$ enabling exponentially many concepts in fewer dimensions.

Under the LRH, concepts can be extracted via \textit{sparse dictionary learning} \citep{olshausenSparseCodingOvercomplete1997}, adapted to neural network activations as \textit{Sparse Autoencoders} (SAEs) \citep{cunninghamSparseAutoencodersFind2023,bricken2023monosemanticity}. An SAE has an encoder $E: \R^m \to \R^C$ and decoder $D: \R^C \to \R^m$ trained via:
\begin{equation}
  \mathcal{L}_{\text{SAE}}(\bx) := \left\Vert
  \psi(x) -
  D(E(\psi(x)))
  \right\Vert_2^2
\end{equation}
With $C \gg m$ (unlike standard autoencoders), sparsity is enforced via L1 penalties or activation functions such as TopK \cite{makhzaniKSparseAutoencoders2014} or BatchTopK \cite{bussmannBatchTopKSparseAutoencoders2024}.

\section{Problem Formulation} \label{sec:lrh_under_symmetries}

To extend the LRH to settings with group symmetries without losing its pragmatic benefits, we reformulate it to take into account how activations transform under input transformations.
Consider a group $G$ acting on inputs $X$, with $g \in G$ and $\bx \in X$.
If $\psi$ is $G$-invariant, $\psi(x) = \psi(g \bx)$ and the same decomposition explains both, but this is not always the case.
This makes interpretation with SAEs harder, as a vanilla SAE would need to disentangle both the feature and the transformation from the activation. For instance, a feature for whether an image contains a specific object can be useful for downstream tasks but the same image would lead to different activations in a model that is not rotation-invariant. To remain faithful to the model, an SAE would need different directions for the same semantic feature, leading to a substantial increase in the number of features and making it harder to identify them.
Many transformations we care about (such as rotations) can be represented as linear transformations. Thus we assume that transformations of model activations can be explained to a large degree by linear transformations; i.e. $\psi(g \bx) \approx \rho(g) \psi(\bx)$ for some $\rho: G \to \R^{m \times m}$. We reformulate the LRH as follows:
\begin{hypothesis}[LRH Under Group Symmetries]
  Let $\psi(\bx) \in \R^m$ be the activations of a neural network at one layer on $\bx \in X \subseteq \R^n$ and $G$ a group that acts on $X$. Then for $g\in G$,
  \begin{align} \label{eq:symmetric_lrh}
    \psi(g \bx) = \rho(g)\psi(\bx) &= \rho(g) \left( \sum_{i=1}^C \bz(x)_i  \bD_i \right) \nonumber \\
    &= \sum_{i=1}^C \bz(x)_i \left( \rho(g) \bD_i \right)
  \end{align}
  for some $\rho: G \to \R^{m \times m}$ with $\bz: \R^n \to \R^C_{\geq 0}$ denoting the strengths of the features $\bD_i \in \R^m$. Similar to the LRH, the support of $\bz(x)$ is sparse.
\end{hypothesis}

\section{Method: Equivariant SAEs} \label{sec:method}

We consider groups $G$ that are products of two cyclic subgroups, i.e. $G = \{ g^p h^q \}_{p,q}$ for two generators $g, h \in G$. This covers cyclic groups, their direct products, and dihedral groups, among others.
Following our extended formulation of the LRH in Equation \ref{eq:symmetric_lrh}, we train our SAEs to learn $G$-invariant latents and hence $G$-invariant reconstructions. We learn a separate linear transformation to model how the activations transform after input transformations.
Thus, with inputs $\bx \in \mathcal{X} \subset \mathbb{R}^n$, activations $\psi(\bx) \in \mathbb{R}^m$, and $p, q = 1,...,\vert G \vert$,
\begin{align}
  D\left(E\left(\psi(g^p h^q \bx)\right)\right) &= \psi(\bx) \\
  {\bM^p \bR^q} D\left(E\left(\psi(g^p h^q \bx)\right)\right) &= \psi({g^p h^q}\bx).
\end{align}
The invariant SAE maps all activations $\psi(g^p \bx)$ of the transformed inputs to $\psi(\bx)$, which is then transformed with $\bM$ and $\bR$ to obtain $\bM^p \bR^q \psi(\bx) = \psi(g^p h^q \bx)$ (Figure \ref{fig:method}, part 3).

\textbf{Loss functions.}
We experiment with three different objectives to make our SAEs approximately invariant to transformations of the base model's inputs:
\begin{align}
  \label{eq:Lc} \mathcal{L}_{\text{Canonical }}(\bx, g) &= \left\Vert \psi(\bx) - D(E(\psi(g \bx))) \right\Vert_2^2  \\
  \label{eq:Ll} \mathcal{L}_{\text{Latent}}(\bx, g) &= \mathcal{L}_{\text{SAE}}(\psi(\bx)) \\
  &\quad + \lambda \left\Vert E(\psi(\bx)) - E(\psi(g \bx)) \right\Vert_2^2 \nonumber \\
  \label{eq:Lo} \mathcal{L}_{\text{Output}}(\bx, g) &= \mathcal{L}_{\text{SAE}}(\psi(\bx)) \\
  &\quad + \lambda \left\Vert D(E(\psi(\bx))) - D(E(\psi(g \bx))) \right\Vert_2^2 \nonumber
\end{align}
where $\bx \in \mathcal{X}$ and $g \in G$. All three of the losses push towards invariant latents but differ in which parts of the SAE are updated with their gradients. %
\begin{itemize}[leftmargin=*]
  \item $\mathcal{L}_{\text{Canonical}}$ \citep{jinLearningGroupActions2024} trains the SAE to reconstruct a canonical sample from each orbit, learning invariance and reconstructions with one term.
  \item $\mathcal{L}_{\text{Output}}$ and $\mathcal{L}_{\text{Latent}}$ separate the two concerns: the reconstruction term $\mathcal{L}_{\text{SAE}}$ is only computed over the canonical inputs, and invariance is imposed directly over SAE latents via $\mathcal{L}_{\text{Latent}}$, or over SAE outputs via $\mathcal{L}_{\text{Output}}$. The core difference is that when optimizing $\mathcal{L}_{\text{Latent}}$, the dictionary is not updated on transformed inputs while $\mathcal{L}_{\text{Output}}$ updates both the encoder and the dictionary.
\end{itemize}
We construct all encoders with two linear layers with a ReLU activation in between. This is motivated by previous work showing that more expressive encoders can lead to more interpretable features \citep{oneillComputeOptimalInference2025}, and the relative simplicity of the encoder ensures that the SAE does not detect features that require complex computations unlikely to be done in a single layer of the base model.

\textbf{Modeling activation-space transformations.}
To map the canonical reconstructions back to their original forms, we need to adapt to the symmetries of the model activations and learn how they transform under input transformations. While equivariant architectures have exact expressions for how their activations are supposed to transform and we can use that in our Equivariant SAEs, such an expression is not generally available. Since many group actions we care about such as rotations can be represented as linear transformations, we have hypothesized that an input-independent linear transformation should be able to explain how model activations transform.

For such a group $G$ with generators $g, h$, we learn two matrices $\bM, \bR \in \mathbb{R}^{m\times m}$ to minimize
\begin{equation}
  \label{eq:L_M}
  \mathcal{L}_M(\bx, p, q) =
  \left\Vert \psi(g^p h^q \bx) - \bM^p \bR^q \psi(\bx) \right\Vert_2^2
\end{equation}
where $\bx \in \mathcal{X}$, $p = 1, ..., \text{ord}(g)$, and $q = 1, ..., \text{ord}(h)$. We initialize $\bM$ and $\bR$ as identity matrices (corresponding to invariant activations) and optimize them using Adam \cite{kingmaAdamMethodStochastic2017}. For cyclic groups, a single matrix suffices. The approach naturally extends to groups requiring more generators by learning additional matrices, though in our experiments we use at most two.

\textbf{Training} (Figure \ref{fig:method}, part 4). Training of our Equivariant SAEs consists of two independent steps that can be done in any order or in parallel. In the first step, the encoder and decoder weights are updated to minimize one of the three invariance objectives, and in the second step, the matrices $\bM$ and if necessary $\bR$ are updated to minimize $\mathcal{L}_M$ (Equation \ref{eq:L_M}). Thus the SAE and the matrices $\bM, \bR$ are trained using different objectives in independent steps. They can be trained over different datasets, e.g. to obtain a domain-specific SAE \citep{muhamedDecodingDarkMatter2025} while $\bM$ and $\bR$ more generally learn how activations transform, although we use the same dataset for both in our experiments.

\section{Experiments}
\label{sec:evaluation}

\begin{figure*}[t!]
  \centering
  \includegraphics[width=\textwidth]{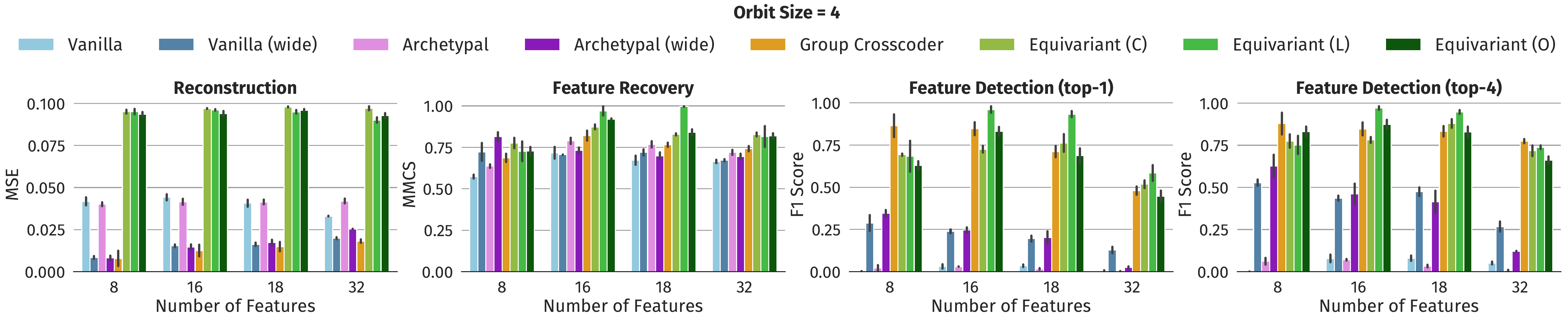}
  \caption{\label{fig:toy_model_metrics}\textbf{Results on our toy model of equivariant features.} Equivariant SAEs (green) outperform the baseline SAEs on feature recovery and detection especially in more challenging cases with more ground truth features and higher interference. Equivariant SAEs lagging behind in reconstruction despite this highlight the importance of evaluations beyond reconstruction quality.}
\end{figure*}

In this section, we describe our experimental results starting with our toy model of equivariant features.
Then we evaluate on a synthetic dataset with geometric shapes as well as real-world galaxy and cell images (samples in Figure \ref{fig:example_images_main} in the Appendix) using 3 types of base models with varying size and complexity: CNNs, MLPs, and Transformers.

Outside of toy settings where ground truth features are known, evaluating interpretability is an open problem. Common approaches such as auto-interp \citep{bills2023language} rely on language models to label features and thus are themselves uninterpretable. Reconstruction quality (MSE between SAE inputs and outputs), while widely used, can be misleading as we demonstrate in our experiments. We therefore take a more pragmatic approach and evaluate the \textit{usefulness} of discovered features via probing \citep{belinkovProbingClassifiersPromises2022}: if SAE features capture meaningful structure, they should support accurate downstream classification of known input properties. This is motivated by findings that SAE probing is a relatively reliable way of comparing SAEs despite SAE evaluations still being an open problem \cite{chanin2026sparse}. We aim to compare different SAEs among themselves, but do not claim that probing over SAE latents/outputs is the optimal approach for such classification tasks, as probing over base model activations often gives better results. All results are averaged over three random seeds and the figures show standard errors.

\textbf{Equivariant SAEs.} We test the 3 versions of our Equivariant SAEs corresponding to the different loss functions described in Section \ref{sec:method}. Each SAE has a two-layer ReLU MLP as its encoder with TopK \citep{makhzaniKSparseAutoencoders2014,gaoScalingEvaluatingSparse2024} output activation.

\textbf{Baselines.} We compare our Equivariant SAEs against 5 baselines: vanilla and Archetypal \citep{felArchetypalSAEAdaptive2025} TopK SAEs, each with a \textit{narrow} and \textit{wide} version, as well as group crosscoders \cite{gortonGroupCrosscodersMechanistic2024}. Archetypal SAEs constrain features to be convex combinations of activation vectors. Narrow SAEs have the same number of latents as the Equivariant SAEs while the wide SAEs have $4\times$ the same number, corresponding to the undesirable scenario of allocating separate latents for each semantic feature in each orientation. All SAEs have the same two-layer encoder module as the Equivariant SAEs (Details in Appendix \ref{app:experimental_details}) and are trained on augmented data. Group crosscoders construct feature vectors with $\vert G \vert$ blocks each to model how features transform under group actions.

\textbf{Evaluation Criteria.} First, on our toy model with known ground truth features, we evaluate how well different SAEs reconstruct their inputs, detect if a feature is active in a given input, and learn a dictionary that aligns with the ground truth features (details below in Section \ref{sec:toy_model_results} and Appendix \ref{app:toy_model_metrics}).
In Sections~\ref{sec:shapes_results}, \ref{sec:galaxy_results}, and \ref{sec:mll_results}, we evaluate the reconstruction quality as well as the feature usefulness of SAEs over actual neural network activations from 3 datasets and 5 base models. We quantify usefulness as the accuracy of an XGBoost \citep{chenXGBoostScalableTree2016} or logistic probing classifier \citep{giulianelliHoodUsingDiagnostic2018,belinkovProbingClassifiersPromises2022} trained on the latents or reconstructions of the SAE to infer various properties of the inputs.

\begin{figure*}[t!]
  \centering
  \includegraphics[width=\textwidth]{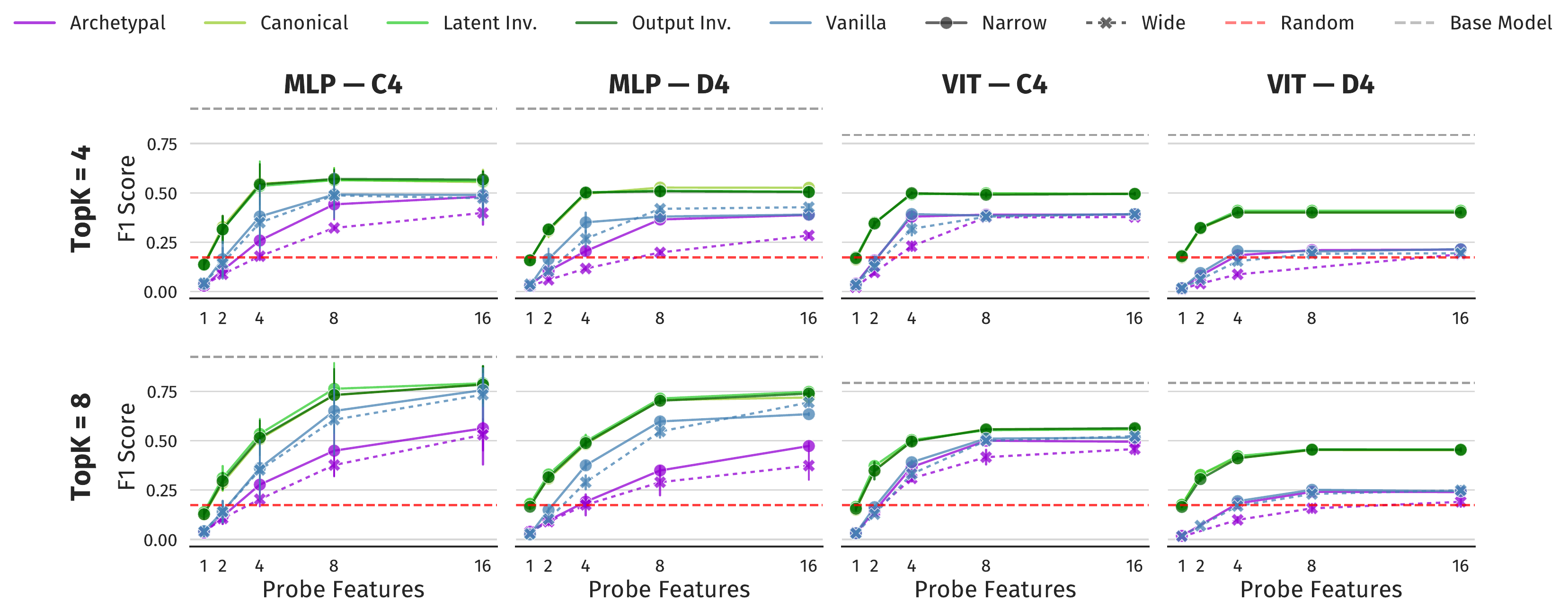}
  \caption{\label{fig:shapes_probing_results}\textbf{Probing SAEs on the Shapes dataset}, averaged over the four task types and three seeds. Equivariant SAEs lead to the most accurate probes across all setups for both C4 and D4 symmetries.}
\end{figure*}

\subsection{Experiment 1: Toy Model} \label{sec:toy_model_results}

Following common practice in mechanistic interpretability research \citep{chaninAbsorptionStudyingFeature2025}, we first design a toy model with known equivariant features (semantically identical but differing in orientation) to understand the behavior of existing SAEs and analyze if an SAE with the correct priors would help alleviate their shortcomings.

The inputs $\bx \in \R^D$ are linear combinations of $N$ (approximately) orthonormal ground truth features $\{ \mathbf{v}_i \}_{i=1}^N$ each independently active with probability $1/N$; i.e. $\mathbf{v}_i^T \mathbf{v}_j \approx 0$ for $i \neq j$ and $\bx = \sum_{i=1}^N \alpha_i \mathbf{v}_i$ with $\alpha_i \sim \text{Ber}(1/N)$. The inputs transform under $\bM \in \R^{D \times D}$ representing the action of a group of order $\vert G \vert$, i.e. $\bM^{\vert G \vert} = \mathbf{I}$.
The ground truth feature activations are invariant with respect to $\bM$, so that $\bM^p \bx = \sum_{i=1}^N \alpha_i \left( \bM^p \mathbf{v}_i \right)$
and only the features transform, which is in line with our extension of the LRH (Equation \ref{eq:symmetric_lrh}).
We fix $D = 16$ and set $N$ to control the degree of \textit{overcompleteness} $N/D$, which determines whether features can be represented orthogonally or must share directions.
\begin{itemize}[leftmargin=*]
  \item Undercomplete ($N=8$): Fewer features than dimensions. Features can occupy nearly orthogonal directions with excess capacity, so no interference is expected.
  \item Complete ($N=16$): Feature count matches dimension and features are orthogonal.
  \item Mildly overcomplete ($N=18$): Slightly more features than dimensions. Exact orthogonality is impossible, leading to weak superposition and low interference.
  \item Strongly overcomplete ($N=32$): Many more features than dimensions. Features must share directions, producing substantial superposition and higher interference.
\end{itemize}

\begin{figure}[t!]
  \centering
  \begin{subfigure}[t]{0.45\textwidth}
    \centering
    \includegraphics[width=0.95\linewidth]{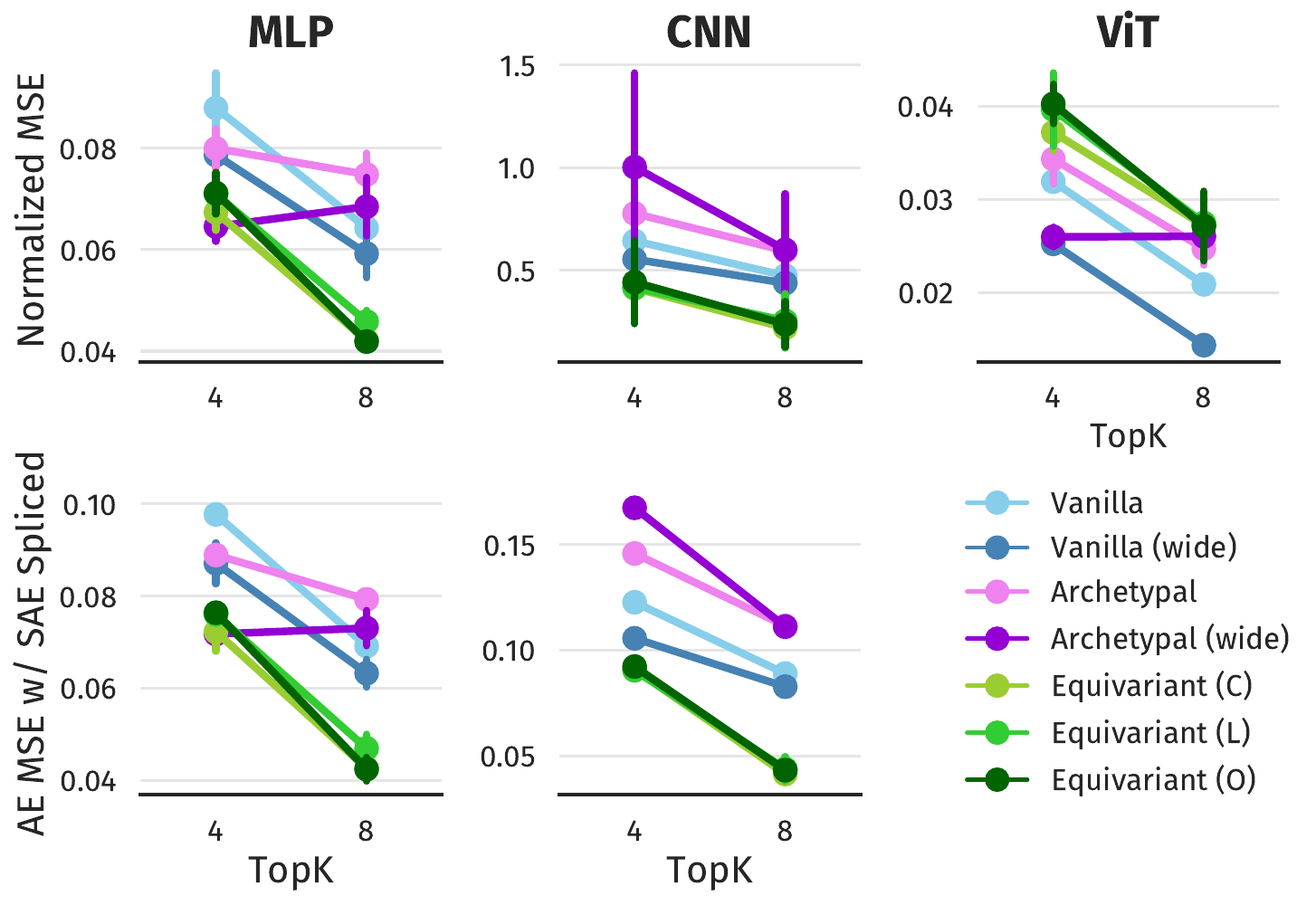}
    \caption{C4}
    \label{fig:shapes_mse_c4}
  \end{subfigure}

  \begin{subfigure}[t]{0.45\textwidth}
    \centering
    \includegraphics[width=0.95\linewidth]{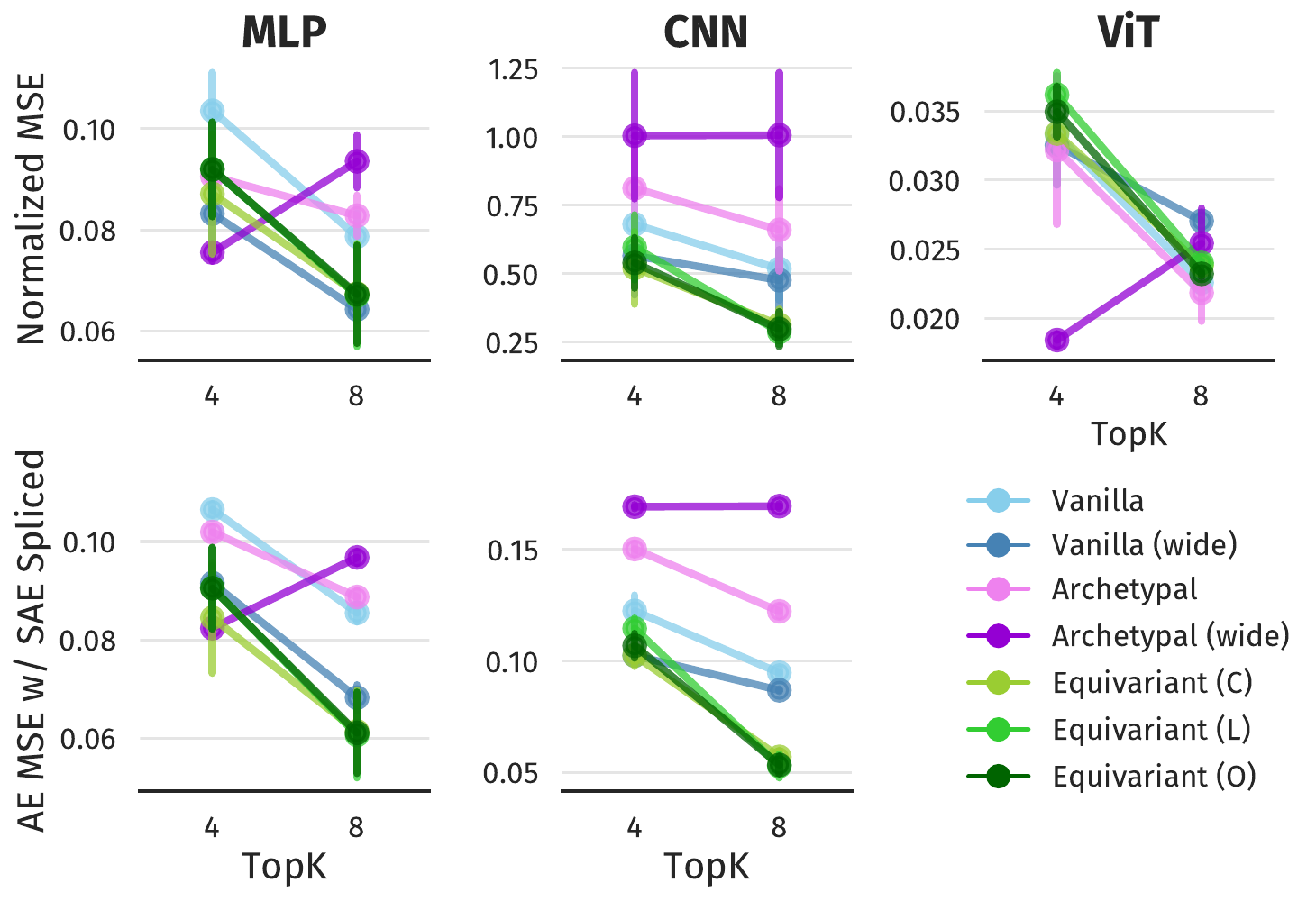}
    \caption{D4}
    \label{fig:shapes_mse_d4}
  \end{subfigure}
  \caption{\label{fig:shapes_mse}\textbf{Reconstruction performance of SAEs on the Shapes dataset}. For both C4 and D4 symmetries, Equivariant SAEs achieve lower reconstruction error than narrow baselines on MLP and CNN autoencoders but perform worse on ViT despite better probing, highlighting that better reconstruction does not always yield more useful features.}
\end{figure}

\textbf{Evaluation Criteria}. We evaluate SAEs on (1) \textit{feature detection}: accuracy of classifying feature activation from top-$m$ candidate SAE latents, (2) \textit{feature recovery}: mean cosine similarity between learned and ground truth features, and (3) \textit{reconstruction error}: MSE between inputs and reconstructions. See Appendix \ref{app:toy_model_metrics} for details.

\textbf{Results.} Figure \ref{fig:toy_model_metrics} displays the results for these criteria for orbit size 4. The results are consistent across orbit sizes and we refer to Appendix \ref{app:further_toy_model} results with orbit sizes 8, 16, and 32. Despite worse reconstructions (first plot), Equivariant SAEs outperform the baseline SAEs on feature recovery and detection as the number of features and thus interference increases. This is surprising, as one would expect given the poorer reconstruction performance of our Equivariant SAEs that they are less interpretable as well. We hypothesize that these results are explained by the non-equivariant SAEs not learning features that are aligned with the data, but instead ``cheating'' by learning features that do not align with any ground truth feature but help reconstruct disparate inputs.
This illustrates the potential risk of relying on reconstruction as the main metric when evaluating interpretability methods and highlights the importance of ensuring that our interpretability methods' priors match what we know about the model and domain we are trying to interpret.

\subsection{Experiment 2: Shapes Dataset}\label{sec:shapes_results}

\begin{itemize}[leftmargin=*, topsep=2pt, itemsep=1pt]
  \item \textbf{Dataset:} We create a synthetic dataset of images with 4 of 8 possible shapes in 4 positions. We consider C4 ($90^\circ$ rotations) and D4 (C4 + vertical flips) symmetries.
  \item \textbf{Base Models:} MLP and CNN autoencoders trained on Shapes dataset; ViT encoder \citep{dosovitskiyImageWorth16x162020} pretrained on ImageNet-1K \citep{dengImageNetLargescaleHierarchical2009}.
  \item \textbf{Probing Tasks:} We construct 180 binary classification tasks. \textbf{S}: Is a particular shape in the image? \textbf{SP}: Is a particular shape in a specific position? \textbf{SO}: Is a particular shape in a specific orientation? \textbf{SPO}: Is a particular shape in a specific position and orientation? Only \textbf{S} is rotation-invariant.
    For each task, we filter SAE latents with the max absolute difference between classes and train XGBoost probes on filtered latents and $\bM$-transformed reconstructions \citep{antvergPitfallsAnalyzingIndividual2022,kantamneniAreSparseAutoencoders2025}.
\end{itemize}

\textbf{Predicting activation transformations}. Table \ref{tab:m_r2} displays the $R^2$ scores between the true and predicted transformed activations of the base models, along with the baseline of setting $\bM = I$. Across all 3 base models (CNN, MLP, ViT), the baseline level of invariance varies between the models, but in all cases, a learned transformation explains more than $97\%$ of the variance in activations resulting from input rotations. These results support our hypothesis that the changes in activations resulting from a group acting linearly on inputs can likewise be explained by linear transformations for both C4 and D4 symmetries.

\textbf{Probing performance.} Figure \ref{fig:shapes_probing_results} displays the F1 scores of probes trained over the truncated reconstructions of the SAEs over the MLP (left) and ViT (right) activations averaged over the four types of probing tasks.
For both TopK values and all truncation lengths, probes trained on Equivariant SAE outputs achieve the best performance under both C4 and D4 symmetries. In particular over the ViT activations under D4 symmetries (right-most plots), non-equivariant SAE probes struggle while Equivariant SAE probes achieve close to 50\% accuracy, showing that accounting for symmetries alone can lead to substantial benefits.

\textbf{Reconstruction quality}. As shown in Figure \ref{fig:shapes_mse}, our Equivariant SAEs tend to outperform the baselines in reconstruction quality for the base MLP and CNN autoencoders, both in terms of the normalized reconstruction MSE, and the base model MSE when the SAE is substituted into the model.
However on the ViT encoder, they lag behind Archetypal and vanilla SAEs despite their better probing performance.
Equivariant SAEs lead to better probing performance relative to the baselines when they are trained on the CNN autoencoder rather than the ViT encoder. The CNN autoencoder is less invariant to C4/D4 transformations (Table \ref{tab:m_r2}). This implies the benefits of Equivariant SAEs are greater when base activations vary more under input rotations.

\begin{table}[t!]
  \centering
  \small
  \caption{\label{tab:m_r2}\textbf{$R^2$ scores for predicting transformed activations}. A learned transformation explains to a large degree how activations transform under C4/D4 symmetries. The benefit is visible even on the relatively more invariant ViT encoder.}
  \begin{tabular}{lllcc}
    \toprule
    \textbf{Dataset} & \textbf{Model} & \textbf{Group} & \textbf{Trained} & \textbf{Id. Baseline} \\
    \midrule
    \multirow{6}{*}{{Shapes}} & \multirow{2}{*}{{CNN}} & C4 & $0.993$ {\tiny\color{gray}$0.0032$} & $0.176$ {\tiny\color{gray}$0.110$} \\
    & & D4 & $0.973$ {\tiny\color{gray}$0.001$} & $0.200$ {\tiny\color{gray}$0.091$} \\
    \cmidrule{2-5}
    & \multirow{2}{*}{{MLP}} & C4 & $0.998$ {\tiny\color{gray}$0.0004$} & $0.668$ {\tiny\color{gray}$0.0195$} \\
    & & D4 & $0.972$ {\tiny\color{gray}$0.011$} & $0.686$ {\tiny\color{gray}$0.018$} \\
    \cmidrule{2-5}
    & \multirow{2}{*}{{ViT}} & C4 & $0.981$ {\tiny\color{gray}$0.0032$} & $0.950$ {\tiny\color{gray}$0.0004$} \\
    & & D4 & $0.984$ {\tiny\color{gray}$0.001$} & $0.950$ {\tiny\color{gray}$0.001$} \\
    \midrule
    \multirow{2}{*}{{Galaxy}} & \multirow{2}{*}{{ZooBot}} & C4 & $0.942$ {\tiny\color{gray}$0.0006$} & $0.908$ {\tiny\color{gray}$0.0009$} \\
    & & D4 & $0.951$ {\tiny\color{gray}$0.0001$} & $0.921$ {\tiny\color{gray}$0.0001$} \\
    \midrule
    \multirow{2}{*}{{MLL23}} & \multirow{2}{*}{{DinoBloom}} & C4 & $0.883$ {\tiny\color{gray}$0.0010$} & $0.777$ {\tiny\color{gray}$0.0009$} \\
    & & D4 & $0.888$ {\tiny\color{gray}$0.0002$} & $0.806$ {\tiny\color{gray}$0.0009$} \\
    \bottomrule
  \end{tabular}
\end{table}

\subsection{Experiment 3: Galaxy Images}\label{sec:galaxy_results}

\begin{itemize}[leftmargin=*, topsep=2pt, itemsep=1pt]
  \item \textbf{Dataset:} GalaxyMNIST \citep{walmsley2022galaxy} with 10,000 galaxy images over 4 invariant classes.
  \item \textbf{Base Model:} ConvNeXT encoder pretrained on 1M galaxy images, fine-tuned on 170k \citep{aussel2025euclid}; SAEs on middle layer activations (channel vectors per pixel).
  \item \textbf{Probing:} 4-way classification on galaxy morphologies; filter SAE latents by highest between-class variance; train logistic probes over average channel vectors per image.
\end{itemize}

\textbf{Predicting activation transformations.} Table \ref{tab:m_r2} shows that training $\bM$ results in $R^2$ scores of 0.94--0.95, an improvement over the baseline ($\bM = \mathbf{I}$) $R^2$ of 0.91--0.92, implying that a learned linear transformation can explain a large part of the variance in activations resulting from input transformations.

\textbf{Probing performance.} Figure \ref{fig:galaxy_probes} (left) displays the F1 scores of the SAE probes as well as base model activations for the GalaxyMNIST classification task. The SAE probes are trained over the SAE latents only since the labels are invariant with respect to C4 and D4. The results are similar to those in Section~\ref{sec:shapes_results}: our Equivariant SAEs result in the most accurate probes across different sparsity budgets (TopK) and as the number of probe features varies, reaching the accuracy of probes trained over the base model activations as the number of features is increased.
The canonical reconstruction (Equation~\ref{eq:Lc}) and output invariance terms (Equation~\ref{eq:Lo}) lead to considerably more accurate probes compared to the latent invariance objective (Equation~\ref{eq:Ll}).

\textbf{Reconstruction quality.} Figure \ref{fig:galaxy_mse} (left) displays the reconstruction MSEs of the SAEs we evaluate on GalaxyMNIST. Despite their superior probing performance, our Equivariant SAEs tend to lag behind vanilla SAEs in reconstruction. Furthermore, probing and reconstruction performances appear inversely correlated: the worst-performing Equivariant SAE for probing, obtained via the latent invariance objective, leads to the best reconstructions among the three Equivariant SAEs. This again highlights that better reconstruction performance does not imply more useful or interpretable features.

\subsection{Experiment 4: Blood Cell Images}\label{sec:mll_results}

\begin{itemize}[leftmargin=*, topsep=2pt, itemsep=1pt]
  \item \textbf{Dataset:} MLL23 \citep{shetab2025large} with 41,906 peripheral blood single-cell images (18 cell types).
  \item \textbf{Base Model:} DinoBloom \citep{koch2024dinobloom}, a DINOv2-based vision transformer trained on over 380k single blood cell images from 13 diverse hematology datasets.
  \item \textbf{Probing:} 18-way cell type classification; filter SAE latents by highest between-class variance; train logistic probes.
\end{itemize}

\textbf{Predicting activation transformations.} Table \ref{tab:m_r2} shows that training $\bM$ results in $R^2$ scores of 0.88--0.89, an improvement over the baseline ($\bM = \mathbf{I}$) $R^2$ of 0.78--0.81, implying that a learned linear transformation can explain a large part of the variance in activations resulting from input transformations.

\begin{figure*}[t!]
  \centering
  \includegraphics[width=\linewidth]{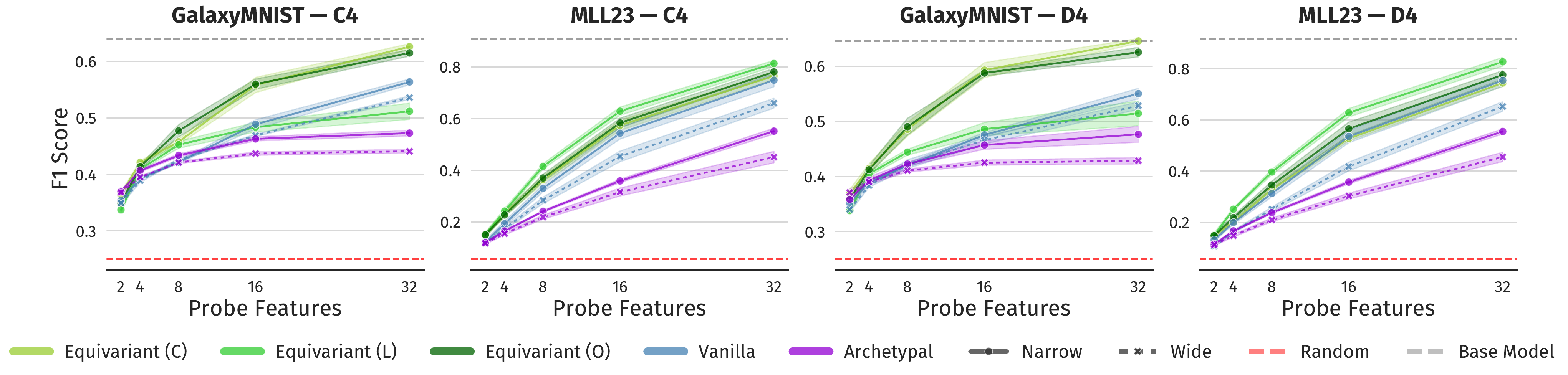}
  \caption{\label{fig:galaxy_probes}\textbf{Probe accuracies on GalaxyMNIST and MLL23} with C4 and D4 symmetries averaged across TopK = 8, 16, 32. Equivariant SAEs lead to better probes, even matching the performance of a probe over the base activations on GalaxyMNIST.}
\end{figure*}

\textbf{Probing performance.} Figure \ref{fig:galaxy_probes} (right) displays the F1 scores of the SAE probes as well as base model activations for the MLL23 classification task. The SAE probes are trained over the SAE latents only since the labels are invariant with respect to C4 and D4. Consistent with the previous experiments, our Equivariant SAEs result in the most accurate probes across different sparsity budgets (TopK) and as the number of probe features varies.
Unlike on GalaxyMNIST, the latent invariance objective (Equation~\ref{eq:Ll}) leads to the best outcomes while the canonical reconstruction (Equation~\ref{eq:Lc}) and output invariance terms (Equation~\ref{eq:Lo}) are less effective.

\begin{figure}[t!]
  \centering
  \includegraphics[width=\linewidth]{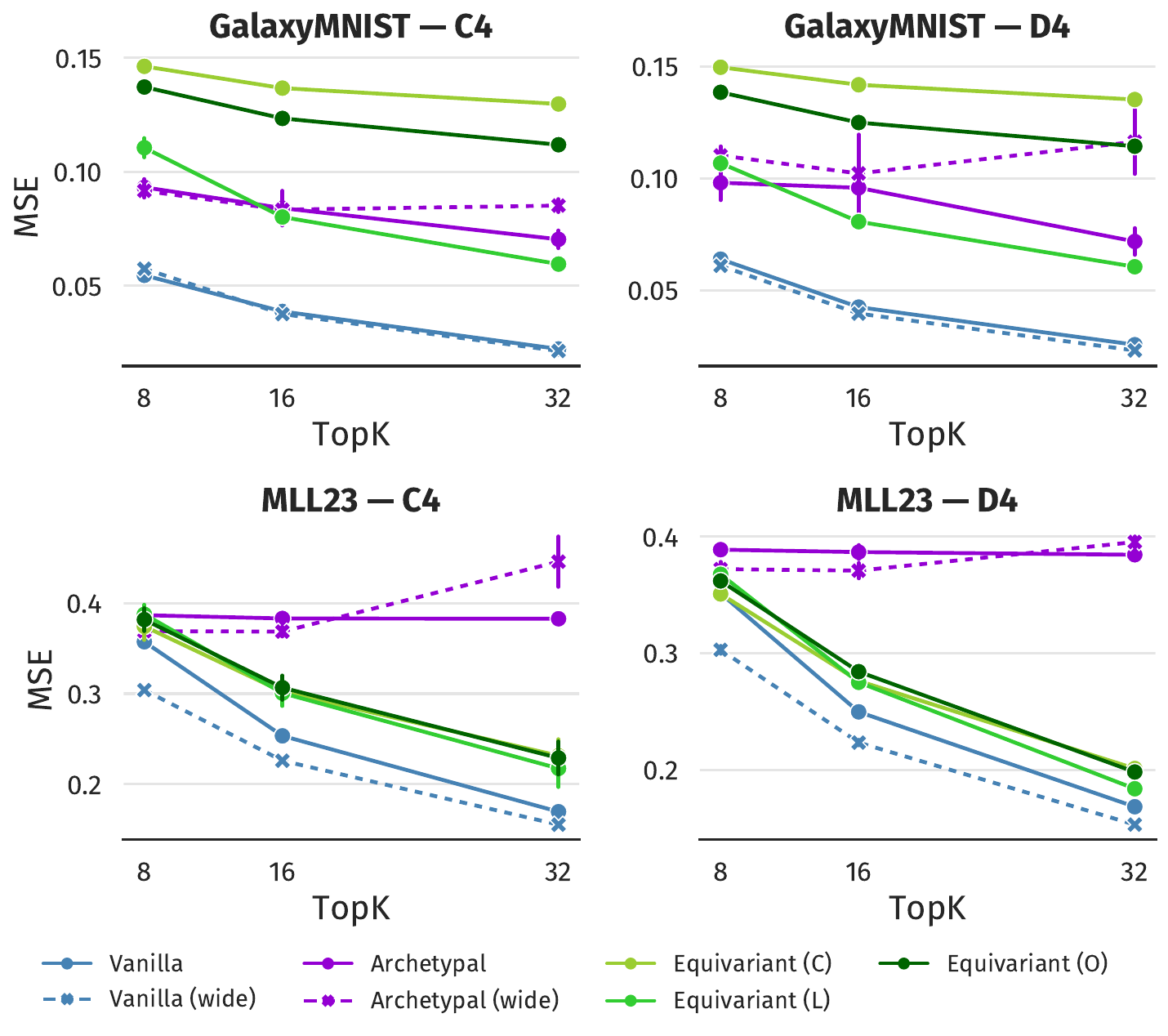}
  \caption{\label{fig:galaxy_mse}\textbf{Reconstruction performances on GalaxyMNIST and MLL23}. The vanilla SAEs have the lowest reconstruction errors while Equivariant SAEs generally perform worse. This is surprising given the superior probing performance of Equivariant SAEs.}
\end{figure}

\textbf{Reconstruction quality.} Figure \ref{fig:galaxy_mse} (right) displays the reconstruction MSEs on MLL23. Similar to GalaxyMNIST, our Equivariant SAEs tend to lag behind vanilla SAEs in reconstruction despite their superior probing performance, further highlighting that better reconstruction does not imply more useful or interpretable features, and that the trade-offs of various methods with reconstruction performance determines the quality of the discovered features.

\section{Related work} \label{sec:related_work}

Our work is one of the first to bridge ideas from the sparse dictionary learning literature and the equivariant representation learning literature with a particular focus on mechanistic interpretability. It relates to several lines of  work.

\textbf{Learning (Approximate) Symmetries.} Objectives to enforce known symmetries similar to our SAE losses have been proposed in \citet{elhagRelaxedEquivarianceMultitask2024,jinLearningGroupActions2024}, and our approach of an invariant encoder/decoder and a separate mapping follows that of \citet{winterUnsupervisedLearningGroup2022}. Our novelty is in our application of these approaches to neural network activations, where the symmetries are induced by the input transformations and may not be well-defined.

\citet{yangLatentSpaceSymmetry2024} and \citet{lairdMatrixNetLearningSymmetry2024} aim to discover symmetries from data. Most relevant for us, \citet{lairdMatrixNetLearningSymmetry2024} propose the MatrixNet architecture to learn matrix representation of group elements while staying faithful to the group axioms.
However, such approaches do not scale well to $O(10^3)$-dimensional activations, requiring orders of magnitudes more parameters to output a transformation matrix. \citet{yangLatentSpaceSymmetry2024} try to avoid this by learning symmetries over a latent space, but this harms interpretability by further distancing the transformations from activation space. In contrast, we learn the transformation over activations.

\textbf{Symmetries and Dictionary Learning}. Group-equivariant sparse dictionary learning methods have been proposed \citep{shewmakeGroupEquivariantSparse2023,shewmakeVisualSceneRepresentation2023} although such exact symmetries cannot be enforced over neural network activations unless the base model is strictly equivariant.   %
\citep{gortonGroupCrosscodersMechanistic2024} proposes group crosscoders to analyze how features learned by a neural network transform as its inputs are transformed by constructing dictionary vectors with $G$ blocks for each feature. The size of the dictionary thus scales linearly with $\vert G \vert$ unlike our approach where the number of parameters is constant with respect to $\vert G \vert$. The group crosscoder furthermore is only trained on canonical inputs, and thus is invariant by construction.

\textbf{Extensions of the LRH and SAE Architectures}. Generalizations of the LRH to structures beyond linear directions have been proposed \citep{engelsNotAllLanguage2025,pearce2025tree}, e.g., by modeling features as manifolds \citep{modellOriginsRepresentationManifolds2025}.
These geometric considerations have in turn informed the design of SAE variants \citep{hindupurProjectingAssumptionsDuality2025,costaFlatHierarchicalExtracting2025}. This line of work broadly aligns with our philosophy of tailoring interpretability tools to how the input data is structured.

Archetypal SAEs \citep{felArchetypalSAEAdaptive2025} model features as convex combinations of activations. Our equivariance constraints however are stricter. As our experiments demonstrate, features that lie in the convex hull of activations may still be uninterpretable. While this does not devalue the Archetypal SAE, it highlights the value of carefully designing what assumptions interpretability tools make.

\section{Conclusion} \label{sec:discussion}

We showed that SAEs can fail on symmetric data in ways difficult to detect through reconstruction metrics, and addressed this by extending the Linear Representation Hypothesis to account for symmetries, guiding the design of \textit{Equivariant SAEs} that decompose activations into invariant features and a learned linear transformation modeling how activations change under input transformations. Our Equivariant SAEs learn features more useful for downstream probing tasks despite lagging behind in reconstruction quality. Key directions for future work include generalizing to larger or continuous groups, leveraging the learned transformation $\bM$ for symmetry-aware feature labeling, and investigating how Equivariant SAE features interact across layers via methods such as Circuit Tracing \citep{ameisen2025circuit}. More broadly, our findings demonstrate that reconstruction quality can be inversely correlated with feature usefulness, and that tailoring the priors of interpretability tools to the structure of the data they operate on can yield significant improvements.

\bibliography{bibliography,manual_bibliography}

\clearpage

\appendix

\section{Reproducibility}

Our experiments are mainly implemented in PyTorch, utilizing existing codebases of baselines (Archetypal SAEs \citep{felArchetypalSAEAdaptive2025}) along with existing mechanistic interpretability tooling (nnsight \citep{fiotto2025nnsight}) and our custom implementations of SAEs and group crosscoders \cite{gortonGroupCrosscodersMechanistic2024} provided as part of the supplementary material. The datasets used in our experiments are either publicly available (GalaxyMNIST \citep{walmsley2022galaxy} and MLL23 \cite{shetab2025large}) or can be generated using our implementation. Each experiment was run on a single NVIDIA RTX 3090 GPU with 24GB of memory.

\section{Toy Model Evaluation Metrics} \label{app:toy_model_metrics}

We evaluate SAEs on our toy model with respect to three metrics:
\begin{itemize}[leftmargin=*]
  \item \textbf{Feature Detection} (top-$m$): For each ground truth feature, we take the $m$ SAE latents with the highest absolute activation difference between the inputs with and without that feature \citep{antvergPitfallsAnalyzingIndividual2022} and train binary logistic probes on those latents to detect if that ground truth feature is active. We report a single F1 score averaged over the ground truth features.

  \item \textbf{Feature Recovery}: Given the SAE dictionary $\mathbf{D}$, we compute the matching $\pi \in S_N$ (permutations) using the Hungarian algorithm \citep{kuhnHungarianMethodAssignment1955} between the SAE features and the ground truth features with the highest average cosine similarity: $\max_{\pi \in S_N} \frac{1}{N}\sum_{i=1}^N \text{CosSim}(\mathbf{D}_i, \mathbf{v}_{\pi(i)})$.

  \item \textbf{Reconstruction}: MSE of the SAE reconstructions.
\end{itemize}

\section{Experimental Details} \label{app:experimental_details}

\subsection{Datasets and Probing Setups}

\subsubsection{Shapes}

\begin{figure}[t]
  \centering
  \includegraphics[width=\linewidth]{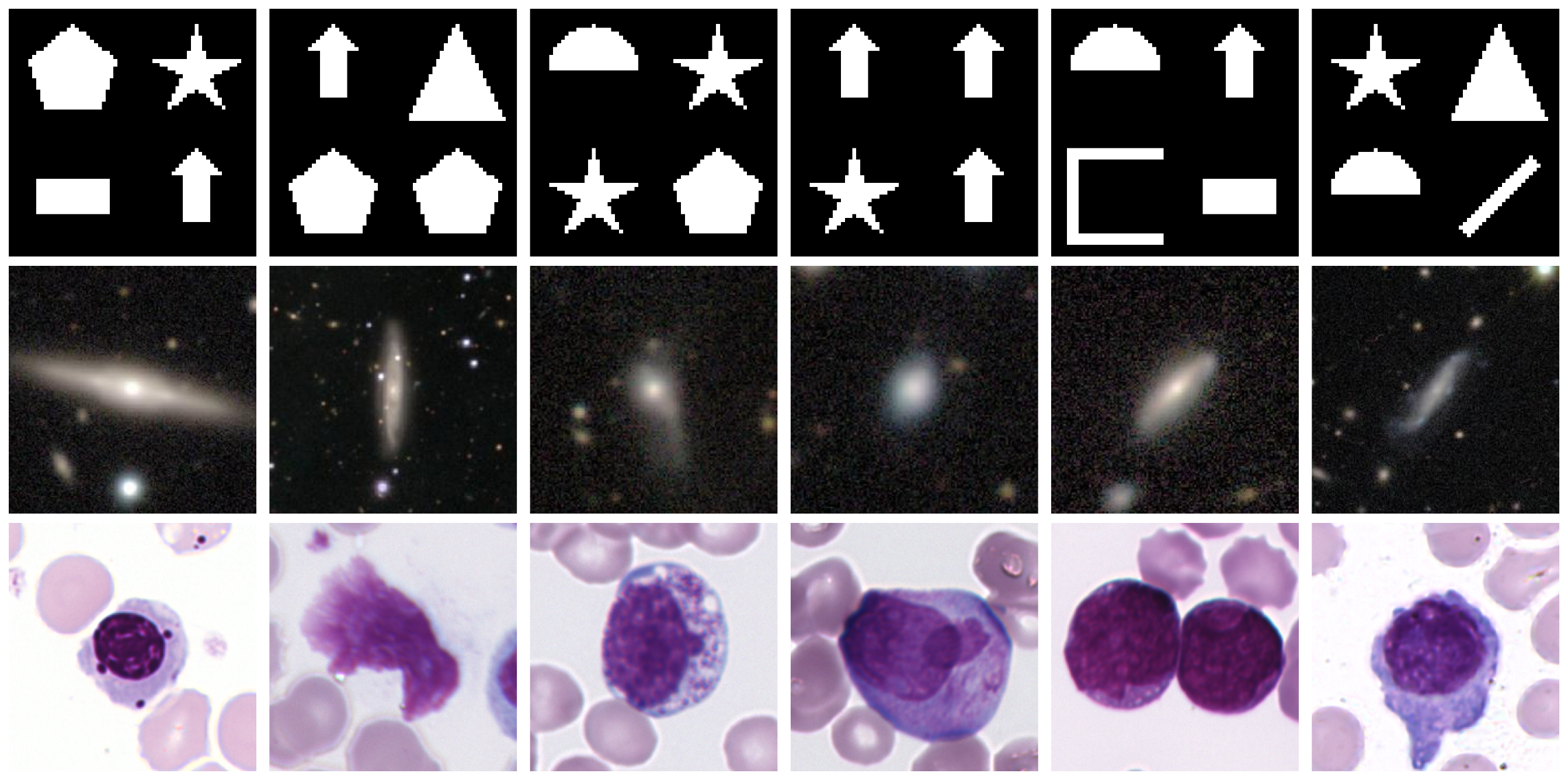}
  \caption{\label{fig:example_images_main}Sample images from the datasets we use in our evaluations. Top to bottom: Shapes, GalaxyMNIST, and MLL23.}
\end{figure}

Figure \ref{fig:example_images_main} displays samples images in the canonical orientation. When rotated in increments of $90^\circ$, the rectangle and the diagonal shapes have two orientations, and the other six shapes have four. Only the rectangle and the "C" shapes are invariant under D4 flips. Each image contains a randomly sampled shape in each of its four quadrants. Precise definitions of our binary probing tasks are then as follows, with a shape's position denoting which of the four quadrants it is in in an image, and its orientation denoting which of the four or two orientations it is in:
\begin{itemize}
  \item \textbf{S}$(s)$: Does the image contain shape $s$ in any position or orientation?
  \item \textbf{SO}$(s,o)$: Does the image contain shape $s$ in orientation $o$ and any position?
  \item \textbf{SP}$(s, p)$: Does the image contain shape $s$ in position $p$ and any orientation?
  \item \textbf{SPO}$(s, p, o)$: Does the image contain shape $s$ in position $p$ and orientation $o$?
\end{itemize}
This results in a total of 8 \textbf{S} (one for each shape), 28 \textbf{SO} (2 shapes $\times$ 2 orientations $+$ 6 shapes $\times$ 4 orientations), 32 \textbf{SP} (8 shapes $\times$ 4 orientations), and 112 \textbf{SPO} (2 shapes $\times$ 2 orientations $\times$ 4 positions $+$ 6 shapes $\times$ 4 orientations $\times$ 4 positions) tasks, for a total of 180 tasks. {\color{CameraReadyUpdate}Note that we report F1 scores rather than raw accuracies as the tasks are not balanced and contain more negative than positive examples, with the share of negative examples ranging from $\sim 58\%$ in the \textbf{S} tasks to $87 - 89 \%$ in the \textbf{SP} and \textbf{SO} tasks, and $97\%$ in the \textbf{SPO} tasks.} The probe we ultimately report the results from, XGBoost \cite{chenXGBoostScalableTree2016}, consists of 100 estimators with a maximum depth of 6, and is trained with the learning rate 0.3 and L2 regularization.

\subsubsection{GalaxyMNIST and MLL23}

GalaxyMNIST \cite{walmsley2022galaxy} and MLL23 \cite{shetab2025large} has 4 and 18 equally distributed C4/D4-invariant labels respectively. Using those labels as targest, we train logistic regression probes using scikit-learn \citep{pedregosaScikitlearnMachineLearning2011}, optimized via L-BFGS \citep{liuLimitedMemoryBFGS1989} with at most 3,000 steps.

\subsection{Base models}

\subsubsection{Shapes}

We train our base autoencoders for 100 epochs over 10,000 randomly generated samples from our dataset augmented with random C4 or D4 transformations depending on the setup with a batch size of 64 using Adam \cite{kingmaAdamMethodStochastic2017} and learning rate 1e-3. Their architectures are detailed in Table \ref{tab:architectures}. The ViT encoder is the \texttt{vit\_b\_16} model sourced from TorchVision \citep{TorchVision_maintainers_and_contributors_TorchVision_PyTorch_s_Computer_2016}. We use the pre-trained weights with no additional training.

\subsubsection{GalaxyMNIST}

The base model for GalaxyMNIST is a ConvNeXT encoder pretrained on 1 million galaxy images and further fine-tuned on an additional 170,000 images \citep{aussel2025euclid}. We train our SAEs over the 256-dimensional middle layer activations, treating each pixel's channel vector as a separate input.

\subsubsection{MLL23}

We use the pre-trained DinoBloom \cite{koch2024dinobloom} model over MLL23. It is a DINOv2-based vision transformer trained on over 380k single blood cell images from 13 datasets. We do not do any additional training, and train the SAEs over the activations of the CLS tokens as they capture the information from the entire image and we are ultimately interested in the usefulness of features in downstream image classification tasks.

\begin{table*}[t!]
  \caption{\label{tab:architectures}\textbf{Architectures of the MLP and CNN autoencoders.} The first section of each model corresponds to the encoder and the second section to the decoder. We train our SAEs over the pre-activation encoder outputs.}
  \centering

  \begin{tabular}{ll}
    \toprule
    \textbf{MLP} & \textbf{CNN} \\
    \midrule
    Input: 4096 (64$\times$64) & Input: 1$\times$64$\times$64 \\
    \midrule
    Linear(4096, 256) & Conv2d(1, 16, 3$\times$3, stride=2, pad=1) \\
    ReLU & ReLU \\
    Linear(256, 256) & Conv2d(16, 32, 3$\times$3, stride=2, pad=1) \\
    ReLU & ReLU \\
    & Conv2d(32, 256, 16$\times$16) \\
    & ReLU \\
    & \\
    Linear(256, 256) & ConvTranspose2d(256, 32, 16$\times$16) \\
    ReLU & ReLU \\
    Linear(256, 4096) & ConvTranspose2d(32, 16, 3$\times$3, stride=2, pad=1, out\_pad=1) \\
    & ReLU \\
    & ConvTranspose2d(16, 1, 3$\times$3, stride=2, pad=1, out\_pad=1) \\
    \bottomrule
  \end{tabular}

\end{table*}

\subsection{Sparse Autoencoders}

All SAEs in our experiments are TopK SAEs with two-layer ReLU MLPs as their encoders and are all trained using Adam \citep{kingmaAdamMethodStochastic2017}, with further training and architecture details as follows:

\textbf{Toy Model}

\begin{itemize}
  \item \textbf{Encoder hidden layer size:} 64
  \item \textbf{Epochs:} 256
  \item \textbf{Learning Rate:} 5e-4
  \item \textbf{TopK:} 3
  \item \textbf{(Equivariant SAE) Invariance coefficient} ($\lambda$): 1
  \item \textbf{(Archetypal SAE) Number of KMeans clusters}: 256
\end{itemize}

\textbf{Shapes Dataset}

\begin{itemize}
  \item \textbf{Encoder hidden layer size: } 128
  \item \textbf{Epochs:} 500
  \item \textbf{Learning Rate:} 5e-4
  \item \textbf{TopK:} 4, 8
  \item \textbf{(Equivariant SAE) Invariance coefficient} ($\lambda$): 1
  \item \textbf{(Archetypal SAE) Number of KMeans clusters}: 2,048
\end{itemize}

\textbf{GalaxyMNIST and MLL23}

\begin{itemize}
  \item \textbf{Encoder hidden layer size: }128
  \item \textbf{Epochs: } 100
  \item \textbf{Learning Rate: } 5e-4
  \item \textbf{TopK:} 8, 16, 32
  \item \textbf{(Equivariant SAE) Invariance coefficient} ($\lambda$): 1
  \item \textbf{(Archetypal SAE) Number of KMeans clusters}: 1,024
\end{itemize}

\section{Further Results on the Toy Model} \label{app:further_toy_model}

\begin{figure*}[t!]
  \centering
  \begin{subfigure}[t]{0.9\textwidth}
    \centering
    \includegraphics[width=\textwidth]{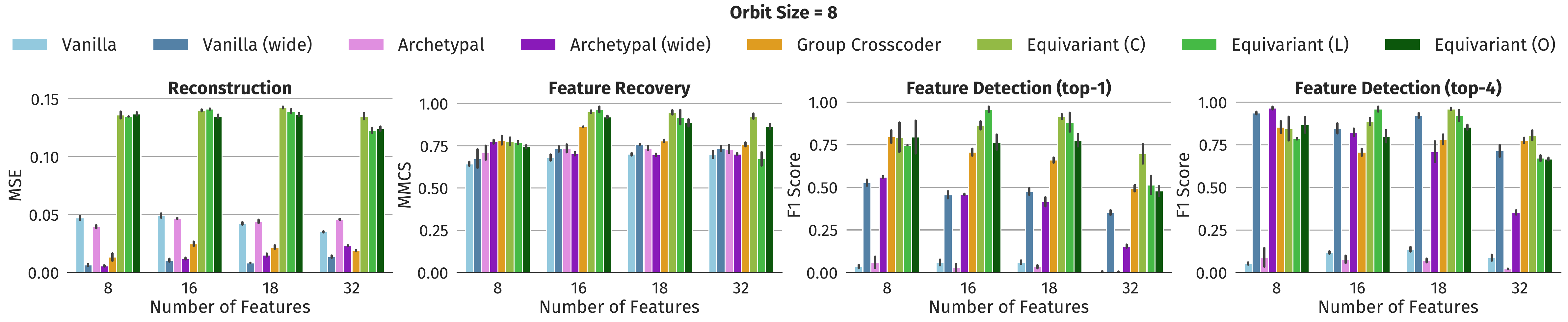}
    \caption{Orbit size 8}
  \end{subfigure}
  \vspace{0.5em}
  \begin{subfigure}[t]{0.9\textwidth}
    \centering
    \includegraphics[width=\textwidth]{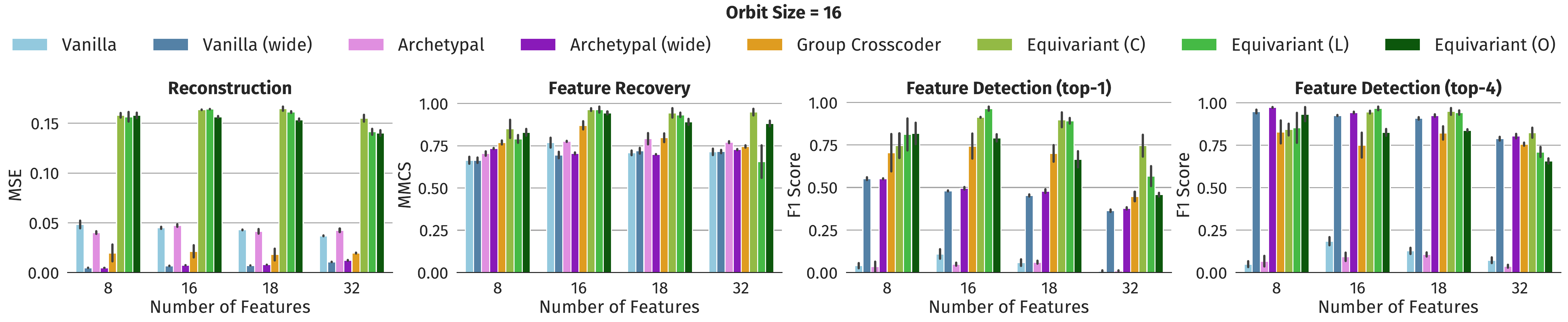}
    \caption{Orbit size 16}
  \end{subfigure}
  \hfill
  \begin{subfigure}[t]{0.9\textwidth}
    \centering
    \includegraphics[width=\textwidth]{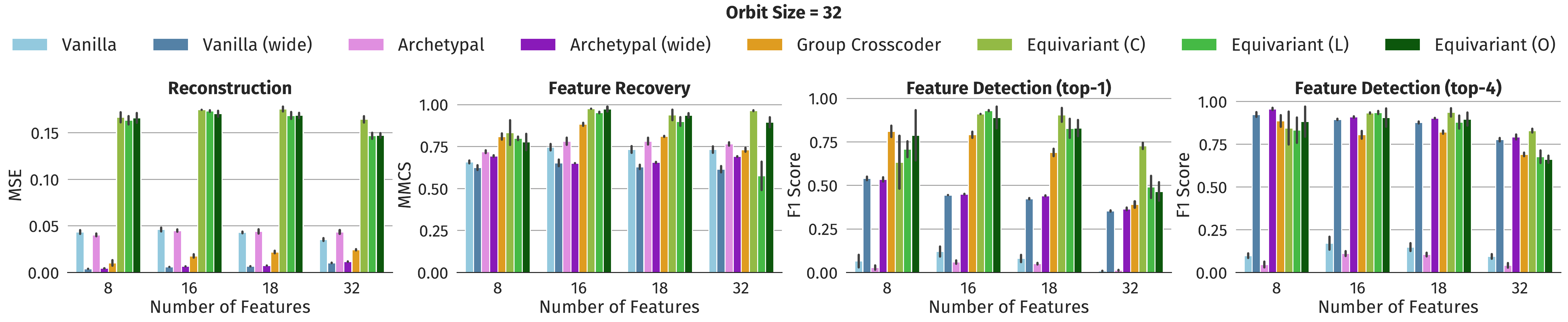}
    \caption{Orbit size 32}
  \end{subfigure}
  \caption{\label{fig:further_toy_model}\textbf{Comparison of various SAEs on our toy model of equivariant features} with orbit sizes 8, 16, and 32. }
\end{figure*}

Figure \ref{fig:further_toy_model} displays the results of the Equivariant and baseline SAEs on our toy model with orbit sizes 8, 16, 32. The results follow a similar trend with those reported in the main body. Despite worse reconstructions, Equivariant SAEs outperform baselines in feature recovery and detection (especially top-1 detection which is more challenging) with significantly fewer parameters than the wide counterparts of the baseline SAEs as well as the group crosscoder.

\section{Further Shapes Probing Results}

Tables \ref{tab:recons_probes_k4} and \ref{tab:recons_probes_k8} display the full probing results over the Shapes dataset averaged over the MLP, CNN, and ViT base models and three seeds. Probes trained over truncated Equivariant SAE outputs consistently outperform those trained over the baseline SAEs' outputs across all tasks. The difference is more significant in the SO, SP, and SPO tasks compared to the S task in which the labels are invariant with respect to C4 and D4, highlighting the increased benfeit of incorporating data symmetries into explanations as those symmetries become more relevant.

\section{Sensitivity to Canonical Orbit Representatives} \label{app:canonical_sensitivity}

In our main experiments, we use a fixed canonical representative for each orbit. Here we investigate the sensitivity of Equivariant SAEs to this choice by re-running the toy model experiments with orbit size 4 and randomly chosen canonical representatives in each run across five seeds. Table~\ref{tab:canonical_sensitivity} reports the results with standard deviations.

While the results vary more compared to the fixed canonical representative setup used in the main paper, the Equivariant SAEs remain accurate in feature recovery and detection despite a slight decrease in performance. This is expected, as choosing different canonical features from different orbits makes the features in the canonical frame less orthogonal to each other and the learning task harder.

\begin{table*}[t!]
  \centering
  \small
  \caption{\label{tab:canonical_sensitivity}\textbf{Toy model results with randomly chosen canonical orbit representatives} (5 seeds). Equivariant SAEs remain effective despite increased variance from random canonical choices.}
  \begin{tabular}{llcccc}
    \toprule
    \textbf{SAE Type} & \textbf{N. Features} & \textbf{Reconstruction} & \textbf{Feature Recovery} & \textbf{Detection (top-1)} & \textbf{Detection (top-4)} \\
    \midrule
    Equivariant (C) & 8  & $0.097 \pm 0.003$ & $0.761 \pm 0.097$ & $0.606 \pm 0.192$ & $0.869 \pm 0.062$ \\
    Equivariant (C) & 16 & $0.097 \pm 0.001$ & $0.844 \pm 0.025$ & $0.620 \pm 0.080$ & $0.837 \pm 0.056$ \\
    Equivariant (C) & 18 & $0.096 \pm 0.001$ & $0.813 \pm 0.036$ & $0.544 \pm 0.097$ & $0.790 \pm 0.020$ \\
    Equivariant (C) & 32 & $0.098 \pm 0.002$ & $0.813 \pm 0.020$ & $0.340 \pm 0.036$ & $0.613 \pm 0.036$ \\
    \midrule
    Equivariant (L) & 8  & $0.096 \pm 0.002$ & $0.825 \pm 0.066$ & $0.636 \pm 0.099$ & $0.930 \pm 0.051$ \\
    Equivariant (L) & 16 & $0.095 \pm 0.001$ & $0.878 \pm 0.055$ & $0.729 \pm 0.119$ & $0.842 \pm 0.081$ \\
    Equivariant (L) & 18 & $0.094 \pm 0.001$ & $0.861 \pm 0.051$ & $0.678 \pm 0.136$ & $0.841 \pm 0.072$ \\
    Equivariant (L) & 32 & $0.093 \pm 0.003$ & $0.771 \pm 0.071$ & $0.445 \pm 0.094$ & $0.630 \pm 0.071$ \\
    \midrule
    Equivariant (O) & 8  & $0.095 \pm 0.002$ & $0.733 \pm 0.090$ & $0.682 \pm 0.101$ & $0.848 \pm 0.109$ \\
    Equivariant (O) & 16 & $0.095 \pm 0.001$ & $0.825 \pm 0.037$ & $0.553 \pm 0.106$ & $0.794 \pm 0.081$ \\
    Equivariant (O) & 18 & $0.094 \pm 0.001$ & $0.790 \pm 0.070$ & $0.498 \pm 0.126$ & $0.723 \pm 0.084$ \\
    Equivariant (O) & 32 & $0.094 \pm 0.002$ & $0.790 \pm 0.035$ & $0.285 \pm 0.043$ & $0.506 \pm 0.049$ \\
    \bottomrule
  \end{tabular}
\end{table*}

\begin{table*}[ht]
\caption{F1 scores on reconstruction probes for $\text{TopK}=4$. Best SAE per group in \textbf{bold}.}
\label{tab:recons_probes_k4}
\centering
\small
\resizebox{\textwidth}{!}{
\begin{tabular}{lllcccc|ccc}
\toprule
\textbf{Trunc.} & \textbf{Sym.} & \textbf{Task} & \textbf{Archetypal} & \textbf{Archetypal (Wide)} & \textbf{Vanilla} & \textbf{Vanilla (Wide)} & \textbf{Canonical} & \textbf{Latent Inv.} & \textbf{Output Inv.} \\
\midrule
\multirow{8}{*}{1} & \multirow{4}{*}{C4} & S & 0.345 {\tiny\textcolor{gray}{0.022}} & 0.157 {\tiny\textcolor{gray}{0.022}} & 0.339 {\tiny\textcolor{gray}{0.022}} & 0.317 {\tiny\textcolor{gray}{0.026}} & \textbf{0.462} {\tiny\textcolor{gray}{0.023}} & 0.419 {\tiny\textcolor{gray}{0.024}} & 0.433 {\tiny\textcolor{gray}{0.024}} \\
 &  & SO & 0.098 {\tiny\textcolor{gray}{0.011}} & 0.050 {\tiny\textcolor{gray}{0.008}} & 0.106 {\tiny\textcolor{gray}{0.012}} & 0.096 {\tiny\textcolor{gray}{0.012}} & 0.450 {\tiny\textcolor{gray}{0.010}} & \textbf{0.454} {\tiny\textcolor{gray}{0.012}} & 0.446 {\tiny\textcolor{gray}{0.012}} \\
 &  & SP & 0.017 {\tiny\textcolor{gray}{0.003}} & 0.015 {\tiny\textcolor{gray}{0.003}} & 0.015 {\tiny\textcolor{gray}{0.003}} & 0.016 {\tiny\textcolor{gray}{0.003}} & 0.075 {\tiny\textcolor{gray}{0.007}} & \textbf{0.081} {\tiny\textcolor{gray}{0.007}} & 0.076 {\tiny\textcolor{gray}{0.007}} \\
 &  & SPO & 0.003 {\tiny\textcolor{gray}{0.001}} & 0.002 {\tiny\textcolor{gray}{0.001}} & 0.008 {\tiny\textcolor{gray}{0.001}} & 0.005 {\tiny\textcolor{gray}{0.001}} & \textbf{0.078} {\tiny\textcolor{gray}{0.005}} & 0.078 {\tiny\textcolor{gray}{0.005}} & 0.078 {\tiny\textcolor{gray}{0.004}} \\
 & \multirow{4}{*}{D4} & S & 0.350 {\tiny\textcolor{gray}{0.018}} & 0.088 {\tiny\textcolor{gray}{0.015}} & 0.355 {\tiny\textcolor{gray}{0.020}} & 0.339 {\tiny\textcolor{gray}{0.023}} & 0.426 {\tiny\textcolor{gray}{0.024}} & 0.441 {\tiny\textcolor{gray}{0.022}} & \textbf{0.443} {\tiny\textcolor{gray}{0.020}} \\
 &  & SO & 0.050 {\tiny\textcolor{gray}{0.005}} & 0.037 {\tiny\textcolor{gray}{0.004}} & 0.039 {\tiny\textcolor{gray}{0.005}} & 0.034 {\tiny\textcolor{gray}{0.004}} & 0.449 {\tiny\textcolor{gray}{0.009}} & \textbf{0.480} {\tiny\textcolor{gray}{0.008}} & 0.461 {\tiny\textcolor{gray}{0.008}} \\
 &  & SP & 0.016 {\tiny\textcolor{gray}{0.003}} & 0.008 {\tiny\textcolor{gray}{0.001}} & 0.022 {\tiny\textcolor{gray}{0.004}} & 0.027 {\tiny\textcolor{gray}{0.006}} & \textbf{0.098} {\tiny\textcolor{gray}{0.008}} & 0.080 {\tiny\textcolor{gray}{0.006}} & 0.091 {\tiny\textcolor{gray}{0.007}} \\
 &  & SPO & 0.002 {\tiny\textcolor{gray}{0.000}} & 0.004 {\tiny\textcolor{gray}{0.001}} & 0.003 {\tiny\textcolor{gray}{0.001}} & 0.006 {\tiny\textcolor{gray}{0.001}} & \textbf{0.104} {\tiny\textcolor{gray}{0.004}} & 0.094 {\tiny\textcolor{gray}{0.003}} & 0.102 {\tiny\textcolor{gray}{0.004}} \\
\midrule
\multirow{8}{*}{2} & \multirow{4}{*}{C4} & S & 0.477 {\tiny\textcolor{gray}{0.023}} & 0.280 {\tiny\textcolor{gray}{0.031}} & 0.503 {\tiny\textcolor{gray}{0.025}} & 0.465 {\tiny\textcolor{gray}{0.022}} & 0.595 {\tiny\textcolor{gray}{0.021}} & \textbf{0.597} {\tiny\textcolor{gray}{0.021}} & 0.596 {\tiny\textcolor{gray}{0.021}} \\
 &  & SO & 0.274 {\tiny\textcolor{gray}{0.013}} & 0.139 {\tiny\textcolor{gray}{0.011}} & 0.303 {\tiny\textcolor{gray}{0.015}} & 0.298 {\tiny\textcolor{gray}{0.014}} & \textbf{0.626} {\tiny\textcolor{gray}{0.010}} & 0.625 {\tiny\textcolor{gray}{0.010}} & 0.619 {\tiny\textcolor{gray}{0.010}} \\
 &  & SP & 0.090 {\tiny\textcolor{gray}{0.006}} & 0.051 {\tiny\textcolor{gray}{0.005}} & 0.111 {\tiny\textcolor{gray}{0.007}} & 0.077 {\tiny\textcolor{gray}{0.006}} & 0.181 {\tiny\textcolor{gray}{0.009}} & \textbf{0.183} {\tiny\textcolor{gray}{0.009}} & 0.172 {\tiny\textcolor{gray}{0.009}} \\
 &  & SPO & 0.078 {\tiny\textcolor{gray}{0.003}} & 0.034 {\tiny\textcolor{gray}{0.002}} & 0.112 {\tiny\textcolor{gray}{0.005}} & 0.076 {\tiny\textcolor{gray}{0.004}} & \textbf{0.260} {\tiny\textcolor{gray}{0.006}} & 0.254 {\tiny\textcolor{gray}{0.006}} & 0.260 {\tiny\textcolor{gray}{0.006}} \\
 & \multirow{4}{*}{D4} & S & 0.488 {\tiny\textcolor{gray}{0.017}} & 0.196 {\tiny\textcolor{gray}{0.021}} & 0.519 {\tiny\textcolor{gray}{0.018}} & 0.500 {\tiny\textcolor{gray}{0.021}} & \textbf{0.599} {\tiny\textcolor{gray}{0.021}} & 0.597 {\tiny\textcolor{gray}{0.020}} & 0.591 {\tiny\textcolor{gray}{0.021}} \\
 &  & SO & 0.191 {\tiny\textcolor{gray}{0.008}} & 0.096 {\tiny\textcolor{gray}{0.007}} & 0.171 {\tiny\textcolor{gray}{0.009}} & 0.109 {\tiny\textcolor{gray}{0.006}} & \textbf{0.612} {\tiny\textcolor{gray}{0.008}} & 0.611 {\tiny\textcolor{gray}{0.008}} & 0.612 {\tiny\textcolor{gray}{0.008}} \\
 &  & SP & 0.081 {\tiny\textcolor{gray}{0.006}} & 0.028 {\tiny\textcolor{gray}{0.002}} & 0.105 {\tiny\textcolor{gray}{0.007}} & 0.081 {\tiny\textcolor{gray}{0.005}} & \textbf{0.199} {\tiny\textcolor{gray}{0.010}} & 0.174 {\tiny\textcolor{gray}{0.010}} & 0.197 {\tiny\textcolor{gray}{0.010}} \\
 &  & SPO & 0.056 {\tiny\textcolor{gray}{0.002}} & 0.018 {\tiny\textcolor{gray}{0.002}} & 0.078 {\tiny\textcolor{gray}{0.003}} & 0.044 {\tiny\textcolor{gray}{0.002}} & 0.261 {\tiny\textcolor{gray}{0.005}} & 0.251 {\tiny\textcolor{gray}{0.005}} & \textbf{0.261} {\tiny\textcolor{gray}{0.005}} \\
\midrule
\multirow{8}{*}{4} & \multirow{4}{*}{C4} & S & 0.610 {\tiny\textcolor{gray}{0.025}} & 0.387 {\tiny\textcolor{gray}{0.035}} & 0.667 {\tiny\textcolor{gray}{0.020}} & 0.627 {\tiny\textcolor{gray}{0.021}} & 0.745 {\tiny\textcolor{gray}{0.018}} & 0.741 {\tiny\textcolor{gray}{0.019}} & \textbf{0.747} {\tiny\textcolor{gray}{0.019}} \\
 &  & SO & 0.510 {\tiny\textcolor{gray}{0.015}} & 0.262 {\tiny\textcolor{gray}{0.016}} & 0.563 {\tiny\textcolor{gray}{0.014}} & 0.464 {\tiny\textcolor{gray}{0.015}} & 0.760 {\tiny\textcolor{gray}{0.009}} & \textbf{0.767} {\tiny\textcolor{gray}{0.009}} & 0.764 {\tiny\textcolor{gray}{0.009}} \\
 &  & SP & 0.212 {\tiny\textcolor{gray}{0.009}} & 0.108 {\tiny\textcolor{gray}{0.007}} & 0.274 {\tiny\textcolor{gray}{0.010}} & 0.222 {\tiny\textcolor{gray}{0.009}} & \textbf{0.405} {\tiny\textcolor{gray}{0.013}} & 0.398 {\tiny\textcolor{gray}{0.013}} & 0.402 {\tiny\textcolor{gray}{0.012}} \\
 &  & SPO & 0.262 {\tiny\textcolor{gray}{0.006}} & 0.103 {\tiny\textcolor{gray}{0.005}} & 0.343 {\tiny\textcolor{gray}{0.006}} & 0.275 {\tiny\textcolor{gray}{0.006}} & 0.509 {\tiny\textcolor{gray}{0.007}} & 0.501 {\tiny\textcolor{gray}{0.007}} & \textbf{0.511} {\tiny\textcolor{gray}{0.007}} \\
 & \multirow{4}{*}{D4} & S & 0.628 {\tiny\textcolor{gray}{0.017}} & 0.284 {\tiny\textcolor{gray}{0.028}} & 0.676 {\tiny\textcolor{gray}{0.019}} & 0.654 {\tiny\textcolor{gray}{0.019}} & 0.752 {\tiny\textcolor{gray}{0.018}} & 0.756 {\tiny\textcolor{gray}{0.019}} & \textbf{0.757} {\tiny\textcolor{gray}{0.018}} \\
 &  & SO & 0.350 {\tiny\textcolor{gray}{0.008}} & 0.149 {\tiny\textcolor{gray}{0.009}} & 0.374 {\tiny\textcolor{gray}{0.010}} & 0.270 {\tiny\textcolor{gray}{0.007}} & 0.714 {\tiny\textcolor{gray}{0.008}} & \textbf{0.731} {\tiny\textcolor{gray}{0.008}} & 0.719 {\tiny\textcolor{gray}{0.008}} \\
 &  & SP & 0.195 {\tiny\textcolor{gray}{0.010}} & 0.072 {\tiny\textcolor{gray}{0.005}} & 0.271 {\tiny\textcolor{gray}{0.011}} & 0.191 {\tiny\textcolor{gray}{0.008}} & \textbf{0.361} {\tiny\textcolor{gray}{0.014}} & 0.358 {\tiny\textcolor{gray}{0.013}} & 0.354 {\tiny\textcolor{gray}{0.014}} \\
 &  & SPO & 0.155 {\tiny\textcolor{gray}{0.003}} & 0.051 {\tiny\textcolor{gray}{0.003}} & 0.212 {\tiny\textcolor{gray}{0.004}} & 0.144 {\tiny\textcolor{gray}{0.003}} & \textbf{0.410} {\tiny\textcolor{gray}{0.006}} & 0.408 {\tiny\textcolor{gray}{0.006}} & 0.408 {\tiny\textcolor{gray}{0.006}} \\
\midrule
\multirow{8}{*}{8} & \multirow{4}{*}{C4} & S & 0.680 {\tiny\textcolor{gray}{0.021}} & 0.469 {\tiny\textcolor{gray}{0.038}} & 0.719 {\tiny\textcolor{gray}{0.018}} & 0.719 {\tiny\textcolor{gray}{0.019}} & 0.770 {\tiny\textcolor{gray}{0.018}} & \textbf{0.781} {\tiny\textcolor{gray}{0.018}} & 0.780 {\tiny\textcolor{gray}{0.018}} \\
 &  & SO & 0.595 {\tiny\textcolor{gray}{0.014}} & 0.392 {\tiny\textcolor{gray}{0.019}} & 0.642 {\tiny\textcolor{gray}{0.011}} & 0.630 {\tiny\textcolor{gray}{0.012}} & 0.793 {\tiny\textcolor{gray}{0.009}} & 0.799 {\tiny\textcolor{gray}{0.009}} & \textbf{0.800} {\tiny\textcolor{gray}{0.009}} \\
 &  & SP & 0.287 {\tiny\textcolor{gray}{0.011}} & 0.190 {\tiny\textcolor{gray}{0.011}} & 0.362 {\tiny\textcolor{gray}{0.012}} & 0.377 {\tiny\textcolor{gray}{0.013}} & 0.465 {\tiny\textcolor{gray}{0.015}} & \textbf{0.465} {\tiny\textcolor{gray}{0.014}} & 0.464 {\tiny\textcolor{gray}{0.014}} \\
 &  & SPO & 0.369 {\tiny\textcolor{gray}{0.006}} & 0.206 {\tiny\textcolor{gray}{0.006}} & 0.439 {\tiny\textcolor{gray}{0.007}} & 0.447 {\tiny\textcolor{gray}{0.007}} & 0.574 {\tiny\textcolor{gray}{0.007}} & \textbf{0.575} {\tiny\textcolor{gray}{0.007}} & 0.569 {\tiny\textcolor{gray}{0.007}} \\
 & \multirow{4}{*}{D4} & S & 0.699 {\tiny\textcolor{gray}{0.016}} & 0.247 {\tiny\textcolor{gray}{0.027}} & 0.715 {\tiny\textcolor{gray}{0.017}} & 0.756 {\tiny\textcolor{gray}{0.015}} & \textbf{0.778} {\tiny\textcolor{gray}{0.019}} & 0.760 {\tiny\textcolor{gray}{0.019}} & 0.773 {\tiny\textcolor{gray}{0.018}} \\
 &  & SO & 0.472 {\tiny\textcolor{gray}{0.009}} & 0.254 {\tiny\textcolor{gray}{0.014}} & 0.427 {\tiny\textcolor{gray}{0.009}} & 0.378 {\tiny\textcolor{gray}{0.008}} & \textbf{0.748} {\tiny\textcolor{gray}{0.007}} & 0.745 {\tiny\textcolor{gray}{0.008}} & 0.742 {\tiny\textcolor{gray}{0.008}} \\
 &  & SP & 0.262 {\tiny\textcolor{gray}{0.010}} & 0.085 {\tiny\textcolor{gray}{0.008}} & 0.339 {\tiny\textcolor{gray}{0.014}} & 0.319 {\tiny\textcolor{gray}{0.011}} & \textbf{0.411} {\tiny\textcolor{gray}{0.014}} & 0.384 {\tiny\textcolor{gray}{0.014}} & 0.396 {\tiny\textcolor{gray}{0.014}} \\
 &  & SPO & 0.244 {\tiny\textcolor{gray}{0.004}} & 0.092 {\tiny\textcolor{gray}{0.005}} & 0.252 {\tiny\textcolor{gray}{0.004}} & 0.224 {\tiny\textcolor{gray}{0.004}} & \textbf{0.450} {\tiny\textcolor{gray}{0.006}} & 0.425 {\tiny\textcolor{gray}{0.006}} & 0.437 {\tiny\textcolor{gray}{0.006}} \\
\midrule
\multirow{8}{*}{16} & \multirow{4}{*}{C4} & S & 0.692 {\tiny\textcolor{gray}{0.020}} & 0.501 {\tiny\textcolor{gray}{0.038}} & 0.711 {\tiny\textcolor{gray}{0.016}} & 0.720 {\tiny\textcolor{gray}{0.019}} & 0.771 {\tiny\textcolor{gray}{0.018}} & \textbf{0.780} {\tiny\textcolor{gray}{0.018}} & 0.776 {\tiny\textcolor{gray}{0.018}} \\
 &  & SO & 0.606 {\tiny\textcolor{gray}{0.013}} & 0.431 {\tiny\textcolor{gray}{0.019}} & 0.647 {\tiny\textcolor{gray}{0.010}} & 0.632 {\tiny\textcolor{gray}{0.012}} & 0.798 {\tiny\textcolor{gray}{0.009}} & \textbf{0.801} {\tiny\textcolor{gray}{0.010}} & 0.800 {\tiny\textcolor{gray}{0.009}} \\
 &  & SP & 0.303 {\tiny\textcolor{gray}{0.011}} & 0.212 {\tiny\textcolor{gray}{0.011}} & 0.381 {\tiny\textcolor{gray}{0.012}} & 0.383 {\tiny\textcolor{gray}{0.013}} & 0.467 {\tiny\textcolor{gray}{0.015}} & 0.472 {\tiny\textcolor{gray}{0.015}} & \textbf{0.472} {\tiny\textcolor{gray}{0.014}} \\
 &  & SPO & 0.382 {\tiny\textcolor{gray}{0.007}} & 0.245 {\tiny\textcolor{gray}{0.007}} & 0.464 {\tiny\textcolor{gray}{0.006}} & 0.446 {\tiny\textcolor{gray}{0.007}} & 0.572 {\tiny\textcolor{gray}{0.007}} & 0.574 {\tiny\textcolor{gray}{0.007}} & \textbf{0.575} {\tiny\textcolor{gray}{0.007}} \\
 & \multirow{4}{*}{D4} & S & 0.717 {\tiny\textcolor{gray}{0.014}} & 0.431 {\tiny\textcolor{gray}{0.036}} & 0.724 {\tiny\textcolor{gray}{0.017}} & 0.764 {\tiny\textcolor{gray}{0.016}} & \textbf{0.779} {\tiny\textcolor{gray}{0.019}} & 0.762 {\tiny\textcolor{gray}{0.019}} & 0.774 {\tiny\textcolor{gray}{0.018}} \\
 &  & SO & 0.478 {\tiny\textcolor{gray}{0.009}} & 0.302 {\tiny\textcolor{gray}{0.011}} & 0.432 {\tiny\textcolor{gray}{0.009}} & 0.383 {\tiny\textcolor{gray}{0.008}} & 0.740 {\tiny\textcolor{gray}{0.008}} & \textbf{0.741} {\tiny\textcolor{gray}{0.008}} & 0.737 {\tiny\textcolor{gray}{0.008}} \\
 &  & SP & 0.281 {\tiny\textcolor{gray}{0.010}} & 0.170 {\tiny\textcolor{gray}{0.010}} & 0.353 {\tiny\textcolor{gray}{0.014}} & 0.337 {\tiny\textcolor{gray}{0.012}} & \textbf{0.410} {\tiny\textcolor{gray}{0.014}} & 0.382 {\tiny\textcolor{gray}{0.014}} & 0.394 {\tiny\textcolor{gray}{0.014}} \\
 &  & SPO & 0.251 {\tiny\textcolor{gray}{0.004}} & 0.141 {\tiny\textcolor{gray}{0.004}} & 0.263 {\tiny\textcolor{gray}{0.005}} & 0.237 {\tiny\textcolor{gray}{0.004}} & \textbf{0.445} {\tiny\textcolor{gray}{0.006}} & 0.424 {\tiny\textcolor{gray}{0.006}} & 0.431 {\tiny\textcolor{gray}{0.006}} \\
\bottomrule
\end{tabular}}
\end{table*}

\begin{table*}[ht]
\caption{F1 scores on reconstruction probes for $\text{TopK}=8$. Best SAE per group in \textbf{bold}.}
\label{tab:recons_probes_k8}
\centering
\small
\resizebox{\textwidth}{!}{
\begin{tabular}{lllcccc|ccc}
\toprule
\textbf{Trunc.} & \textbf{Sym.} & \textbf{Task} & \textbf{Archetypal} & \textbf{Archetypal (Wide)} & \textbf{Vanilla} & \textbf{Vanilla (Wide)} & \textbf{Canonical} & \textbf{Latent Inv.} & \textbf{Output Inv.} \\
\midrule
\multirow{8}{*}{1} & \multirow{4}{*}{C4} & S & 0.357 {\tiny\textcolor{gray}{0.020}} & 0.320 {\tiny\textcolor{gray}{0.027}} & 0.343 {\tiny\textcolor{gray}{0.018}} & 0.331 {\tiny\textcolor{gray}{0.023}} & 0.432 {\tiny\textcolor{gray}{0.019}} & \textbf{0.465} {\tiny\textcolor{gray}{0.018}} & 0.433 {\tiny\textcolor{gray}{0.018}} \\
 &  & SO & 0.071 {\tiny\textcolor{gray}{0.009}} & 0.082 {\tiny\textcolor{gray}{0.010}} & 0.081 {\tiny\textcolor{gray}{0.011}} & 0.094 {\tiny\textcolor{gray}{0.011}} & 0.450 {\tiny\textcolor{gray}{0.011}} & \textbf{0.465} {\tiny\textcolor{gray}{0.010}} & 0.441 {\tiny\textcolor{gray}{0.012}} \\
 &  & SP & 0.022 {\tiny\textcolor{gray}{0.004}} & 0.020 {\tiny\textcolor{gray}{0.003}} & 0.017 {\tiny\textcolor{gray}{0.003}} & 0.022 {\tiny\textcolor{gray}{0.005}} & \textbf{0.112} {\tiny\textcolor{gray}{0.008}} & 0.083 {\tiny\textcolor{gray}{0.007}} & 0.101 {\tiny\textcolor{gray}{0.008}} \\
 &  & SPO & 0.005 {\tiny\textcolor{gray}{0.001}} & 0.006 {\tiny\textcolor{gray}{0.001}} & 0.007 {\tiny\textcolor{gray}{0.001}} & 0.008 {\tiny\textcolor{gray}{0.002}} & 0.096 {\tiny\textcolor{gray}{0.006}} & 0.094 {\tiny\textcolor{gray}{0.006}} & \textbf{0.103} {\tiny\textcolor{gray}{0.006}} \\
 & \multirow{4}{*}{D4} & S & 0.395 {\tiny\textcolor{gray}{0.020}} & 0.205 {\tiny\textcolor{gray}{0.028}} & 0.326 {\tiny\textcolor{gray}{0.019}} & 0.366 {\tiny\textcolor{gray}{0.022}} & 0.460 {\tiny\textcolor{gray}{0.022}} & \textbf{0.489} {\tiny\textcolor{gray}{0.019}} & 0.482 {\tiny\textcolor{gray}{0.021}} \\
 &  & SO & 0.058 {\tiny\textcolor{gray}{0.007}} & 0.044 {\tiny\textcolor{gray}{0.006}} & 0.045 {\tiny\textcolor{gray}{0.007}} & 0.037 {\tiny\textcolor{gray}{0.005}} & \textbf{0.475} {\tiny\textcolor{gray}{0.008}} & 0.471 {\tiny\textcolor{gray}{0.008}} & 0.462 {\tiny\textcolor{gray}{0.008}} \\
 &  & SP & 0.026 {\tiny\textcolor{gray}{0.005}} & 0.010 {\tiny\textcolor{gray}{0.002}} & 0.031 {\tiny\textcolor{gray}{0.009}} & 0.040 {\tiny\textcolor{gray}{0.009}} & 0.126 {\tiny\textcolor{gray}{0.009}} & \textbf{0.136} {\tiny\textcolor{gray}{0.009}} & 0.126 {\tiny\textcolor{gray}{0.009}} \\
 &  & SPO & 0.002 {\tiny\textcolor{gray}{0.000}} & 0.004 {\tiny\textcolor{gray}{0.001}} & 0.004 {\tiny\textcolor{gray}{0.001}} & 0.006 {\tiny\textcolor{gray}{0.001}} & \textbf{0.128} {\tiny\textcolor{gray}{0.005}} & 0.120 {\tiny\textcolor{gray}{0.005}} & 0.117 {\tiny\textcolor{gray}{0.005}} \\
\midrule
\multirow{8}{*}{2} & \multirow{4}{*}{C4} & S & 0.483 {\tiny\textcolor{gray}{0.017}} & 0.465 {\tiny\textcolor{gray}{0.024}} & 0.517 {\tiny\textcolor{gray}{0.021}} & 0.497 {\tiny\textcolor{gray}{0.018}} & 0.600 {\tiny\textcolor{gray}{0.020}} & \textbf{0.622} {\tiny\textcolor{gray}{0.020}} & 0.613 {\tiny\textcolor{gray}{0.020}} \\
 &  & SO & 0.251 {\tiny\textcolor{gray}{0.014}} & 0.255 {\tiny\textcolor{gray}{0.013}} & 0.311 {\tiny\textcolor{gray}{0.013}} & 0.297 {\tiny\textcolor{gray}{0.014}} & 0.630 {\tiny\textcolor{gray}{0.009}} & \textbf{0.643} {\tiny\textcolor{gray}{0.010}} & 0.634 {\tiny\textcolor{gray}{0.009}} \\
 &  & SP & 0.090 {\tiny\textcolor{gray}{0.006}} & 0.087 {\tiny\textcolor{gray}{0.006}} & 0.122 {\tiny\textcolor{gray}{0.007}} & 0.117 {\tiny\textcolor{gray}{0.008}} & 0.243 {\tiny\textcolor{gray}{0.011}} & 0.234 {\tiny\textcolor{gray}{0.010}} & \textbf{0.248} {\tiny\textcolor{gray}{0.011}} \\
 &  & SPO & 0.086 {\tiny\textcolor{gray}{0.004}} & 0.078 {\tiny\textcolor{gray}{0.004}} & 0.122 {\tiny\textcolor{gray}{0.004}} & 0.108 {\tiny\textcolor{gray}{0.004}} & 0.309 {\tiny\textcolor{gray}{0.008}} & \textbf{0.315} {\tiny\textcolor{gray}{0.007}} & 0.302 {\tiny\textcolor{gray}{0.007}} \\
 & \multirow{4}{*}{D4} & S & 0.453 {\tiny\textcolor{gray}{0.016}} & 0.148 {\tiny\textcolor{gray}{0.024}} & 0.459 {\tiny\textcolor{gray}{0.013}} & 0.500 {\tiny\textcolor{gray}{0.022}} & \textbf{0.621} {\tiny\textcolor{gray}{0.020}} & 0.612 {\tiny\textcolor{gray}{0.020}} & 0.617 {\tiny\textcolor{gray}{0.020}} \\
 &  & SO & 0.137 {\tiny\textcolor{gray}{0.010}} & 0.119 {\tiny\textcolor{gray}{0.011}} & 0.222 {\tiny\textcolor{gray}{0.011}} & 0.125 {\tiny\textcolor{gray}{0.007}} & 0.612 {\tiny\textcolor{gray}{0.008}} & \textbf{0.624} {\tiny\textcolor{gray}{0.007}} & 0.619 {\tiny\textcolor{gray}{0.007}} \\
 &  & SP & 0.106 {\tiny\textcolor{gray}{0.014}} & 0.027 {\tiny\textcolor{gray}{0.006}} & 0.125 {\tiny\textcolor{gray}{0.014}} & 0.125 {\tiny\textcolor{gray}{0.012}} & 0.217 {\tiny\textcolor{gray}{0.010}} & \textbf{0.228} {\tiny\textcolor{gray}{0.010}} & 0.225 {\tiny\textcolor{gray}{0.010}} \\
 &  & SPO & 0.054 {\tiny\textcolor{gray}{0.003}} & 0.025 {\tiny\textcolor{gray}{0.003}} & 0.086 {\tiny\textcolor{gray}{0.004}} & 0.065 {\tiny\textcolor{gray}{0.003}} & 0.271 {\tiny\textcolor{gray}{0.005}} & \textbf{0.285} {\tiny\textcolor{gray}{0.006}} & 0.270 {\tiny\textcolor{gray}{0.006}} \\
\midrule
\multirow{8}{*}{4} & \multirow{4}{*}{C4} & S & 0.624 {\tiny\textcolor{gray}{0.021}} & 0.594 {\tiny\textcolor{gray}{0.025}} & 0.664 {\tiny\textcolor{gray}{0.019}} & 0.659 {\tiny\textcolor{gray}{0.020}} & 0.734 {\tiny\textcolor{gray}{0.018}} & \textbf{0.752} {\tiny\textcolor{gray}{0.018}} & 0.739 {\tiny\textcolor{gray}{0.018}} \\
 &  & SO & 0.450 {\tiny\textcolor{gray}{0.016}} & 0.447 {\tiny\textcolor{gray}{0.017}} & 0.551 {\tiny\textcolor{gray}{0.013}} & 0.519 {\tiny\textcolor{gray}{0.013}} & 0.767 {\tiny\textcolor{gray}{0.009}} & 0.765 {\tiny\textcolor{gray}{0.009}} & \textbf{0.768} {\tiny\textcolor{gray}{0.009}} \\
 &  & SP & 0.219 {\tiny\textcolor{gray}{0.009}} & 0.185 {\tiny\textcolor{gray}{0.008}} & 0.289 {\tiny\textcolor{gray}{0.010}} & 0.283 {\tiny\textcolor{gray}{0.011}} & \textbf{0.430} {\tiny\textcolor{gray}{0.013}} & 0.411 {\tiny\textcolor{gray}{0.012}} & 0.425 {\tiny\textcolor{gray}{0.012}} \\
 &  & SPO & 0.255 {\tiny\textcolor{gray}{0.006}} & 0.206 {\tiny\textcolor{gray}{0.006}} & 0.351 {\tiny\textcolor{gray}{0.006}} & 0.317 {\tiny\textcolor{gray}{0.006}} & 0.507 {\tiny\textcolor{gray}{0.007}} & \textbf{0.515} {\tiny\textcolor{gray}{0.007}} & 0.508 {\tiny\textcolor{gray}{0.007}} \\
 & \multirow{4}{*}{D4} & S & 0.619 {\tiny\textcolor{gray}{0.021}} & 0.349 {\tiny\textcolor{gray}{0.035}} & 0.692 {\tiny\textcolor{gray}{0.017}} & 0.661 {\tiny\textcolor{gray}{0.019}} & \textbf{0.762} {\tiny\textcolor{gray}{0.020}} & 0.746 {\tiny\textcolor{gray}{0.020}} & 0.747 {\tiny\textcolor{gray}{0.020}} \\
 &  & SO & 0.262 {\tiny\textcolor{gray}{0.009}} & 0.167 {\tiny\textcolor{gray}{0.010}} & 0.376 {\tiny\textcolor{gray}{0.010}} & 0.330 {\tiny\textcolor{gray}{0.009}} & 0.731 {\tiny\textcolor{gray}{0.008}} & \textbf{0.740} {\tiny\textcolor{gray}{0.008}} & 0.730 {\tiny\textcolor{gray}{0.008}} \\
 &  & SP & 0.210 {\tiny\textcolor{gray}{0.012}} & 0.076 {\tiny\textcolor{gray}{0.008}} & 0.278 {\tiny\textcolor{gray}{0.013}} & 0.297 {\tiny\textcolor{gray}{0.015}} & 0.378 {\tiny\textcolor{gray}{0.013}} & 0.392 {\tiny\textcolor{gray}{0.013}} & \textbf{0.395} {\tiny\textcolor{gray}{0.013}} \\
 &  & SPO & 0.141 {\tiny\textcolor{gray}{0.003}} & 0.065 {\tiny\textcolor{gray}{0.003}} & 0.229 {\tiny\textcolor{gray}{0.005}} & 0.196 {\tiny\textcolor{gray}{0.005}} & 0.423 {\tiny\textcolor{gray}{0.006}} & \textbf{0.436} {\tiny\textcolor{gray}{0.006}} & 0.434 {\tiny\textcolor{gray}{0.006}} \\
\midrule
\multirow{8}{*}{8} & \multirow{4}{*}{C4} & S & 0.705 {\tiny\textcolor{gray}{0.023}} & 0.673 {\tiny\textcolor{gray}{0.025}} & 0.779 {\tiny\textcolor{gray}{0.014}} & 0.773 {\tiny\textcolor{gray}{0.018}} & 0.835 {\tiny\textcolor{gray}{0.016}} & 0.835 {\tiny\textcolor{gray}{0.018}} & \textbf{0.840} {\tiny\textcolor{gray}{0.017}} \\
 &  & SO & 0.628 {\tiny\textcolor{gray}{0.015}} & 0.599 {\tiny\textcolor{gray}{0.016}} & 0.747 {\tiny\textcolor{gray}{0.009}} & 0.728 {\tiny\textcolor{gray}{0.010}} & 0.863 {\tiny\textcolor{gray}{0.008}} & \textbf{0.865} {\tiny\textcolor{gray}{0.008}} & 0.865 {\tiny\textcolor{gray}{0.008}} \\
 &  & SP & 0.344 {\tiny\textcolor{gray}{0.012}} & 0.329 {\tiny\textcolor{gray}{0.013}} & 0.539 {\tiny\textcolor{gray}{0.011}} & 0.494 {\tiny\textcolor{gray}{0.013}} & 0.609 {\tiny\textcolor{gray}{0.014}} & \textbf{0.615} {\tiny\textcolor{gray}{0.015}} & 0.610 {\tiny\textcolor{gray}{0.014}} \\
 &  & SPO & 0.438 {\tiny\textcolor{gray}{0.007}} & 0.385 {\tiny\textcolor{gray}{0.008}} & 0.602 {\tiny\textcolor{gray}{0.006}} & 0.563 {\tiny\textcolor{gray}{0.007}} & 0.696 {\tiny\textcolor{gray}{0.007}} & \textbf{0.699} {\tiny\textcolor{gray}{0.007}} & 0.691 {\tiny\textcolor{gray}{0.007}} \\
 & \multirow{4}{*}{D4} & S & 0.741 {\tiny\textcolor{gray}{0.021}} & 0.426 {\tiny\textcolor{gray}{0.040}} & 0.805 {\tiny\textcolor{gray}{0.014}} & 0.780 {\tiny\textcolor{gray}{0.017}} & \textbf{0.849} {\tiny\textcolor{gray}{0.018}} & 0.840 {\tiny\textcolor{gray}{0.017}} & 0.840 {\tiny\textcolor{gray}{0.018}} \\
 &  & SO & 0.447 {\tiny\textcolor{gray}{0.008}} & 0.243 {\tiny\textcolor{gray}{0.012}} & 0.521 {\tiny\textcolor{gray}{0.011}} & 0.485 {\tiny\textcolor{gray}{0.010}} & 0.810 {\tiny\textcolor{gray}{0.008}} & \textbf{0.821} {\tiny\textcolor{gray}{0.007}} & 0.817 {\tiny\textcolor{gray}{0.007}} \\
 &  & SP & 0.398 {\tiny\textcolor{gray}{0.016}} & 0.130 {\tiny\textcolor{gray}{0.010}} & 0.507 {\tiny\textcolor{gray}{0.016}} & 0.494 {\tiny\textcolor{gray}{0.016}} & 0.531 {\tiny\textcolor{gray}{0.016}} & \textbf{0.542} {\tiny\textcolor{gray}{0.016}} & 0.531 {\tiny\textcolor{gray}{0.016}} \\
 &  & SPO & 0.275 {\tiny\textcolor{gray}{0.005}} & 0.118 {\tiny\textcolor{gray}{0.004}} & 0.373 {\tiny\textcolor{gray}{0.006}} & 0.353 {\tiny\textcolor{gray}{0.006}} & 0.571 {\tiny\textcolor{gray}{0.007}} & 0.574 {\tiny\textcolor{gray}{0.007}} & \textbf{0.578} {\tiny\textcolor{gray}{0.007}} \\
\midrule
\multirow{8}{*}{16} & \multirow{4}{*}{C4} & S & 0.759 {\tiny\textcolor{gray}{0.020}} & 0.734 {\tiny\textcolor{gray}{0.022}} & 0.824 {\tiny\textcolor{gray}{0.013}} & 0.829 {\tiny\textcolor{gray}{0.015}} & 0.869 {\tiny\textcolor{gray}{0.015}} & 0.868 {\tiny\textcolor{gray}{0.015}} & \textbf{0.872} {\tiny\textcolor{gray}{0.015}} \\
 &  & SO & 0.700 {\tiny\textcolor{gray}{0.014}} & 0.670 {\tiny\textcolor{gray}{0.015}} & 0.806 {\tiny\textcolor{gray}{0.008}} & 0.806 {\tiny\textcolor{gray}{0.009}} & 0.891 {\tiny\textcolor{gray}{0.007}} & 0.889 {\tiny\textcolor{gray}{0.008}} & \textbf{0.891} {\tiny\textcolor{gray}{0.008}} \\
 &  & SP & 0.437 {\tiny\textcolor{gray}{0.014}} & 0.406 {\tiny\textcolor{gray}{0.014}} & 0.629 {\tiny\textcolor{gray}{0.012}} & 0.609 {\tiny\textcolor{gray}{0.014}} & 0.665 {\tiny\textcolor{gray}{0.016}} & 0.659 {\tiny\textcolor{gray}{0.015}} & \textbf{0.665} {\tiny\textcolor{gray}{0.015}} \\
 &  & SPO & 0.517 {\tiny\textcolor{gray}{0.007}} & 0.481 {\tiny\textcolor{gray}{0.008}} & 0.689 {\tiny\textcolor{gray}{0.006}} & 0.662 {\tiny\textcolor{gray}{0.007}} & \textbf{0.736} {\tiny\textcolor{gray}{0.007}} & 0.731 {\tiny\textcolor{gray}{0.007}} & 0.735 {\tiny\textcolor{gray}{0.007}} \\
 & \multirow{4}{*}{D4} & S & 0.787 {\tiny\textcolor{gray}{0.018}} & 0.457 {\tiny\textcolor{gray}{0.042}} & 0.836 {\tiny\textcolor{gray}{0.015}} & 0.849 {\tiny\textcolor{gray}{0.015}} & 0.857 {\tiny\textcolor{gray}{0.018}} & 0.859 {\tiny\textcolor{gray}{0.017}} & \textbf{0.861} {\tiny\textcolor{gray}{0.017}} \\
 &  & SO & 0.507 {\tiny\textcolor{gray}{0.009}} & 0.291 {\tiny\textcolor{gray}{0.014}} & 0.551 {\tiny\textcolor{gray}{0.010}} & 0.553 {\tiny\textcolor{gray}{0.011}} & 0.824 {\tiny\textcolor{gray}{0.007}} & \textbf{0.836} {\tiny\textcolor{gray}{0.007}} & 0.831 {\tiny\textcolor{gray}{0.007}} \\
 &  & SP & 0.454 {\tiny\textcolor{gray}{0.015}} & 0.161 {\tiny\textcolor{gray}{0.011}} & 0.570 {\tiny\textcolor{gray}{0.017}} & \textbf{0.639} {\tiny\textcolor{gray}{0.016}} & 0.582 {\tiny\textcolor{gray}{0.017}} & 0.599 {\tiny\textcolor{gray}{0.017}} & 0.596 {\tiny\textcolor{gray}{0.017}} \\
 &  & SPO & 0.315 {\tiny\textcolor{gray}{0.005}} & 0.156 {\tiny\textcolor{gray}{0.005}} & 0.406 {\tiny\textcolor{gray}{0.006}} & 0.445 {\tiny\textcolor{gray}{0.007}} & 0.597 {\tiny\textcolor{gray}{0.007}} & \textbf{0.616} {\tiny\textcolor{gray}{0.007}} & 0.611 {\tiny\textcolor{gray}{0.007}} \\
\bottomrule
\end{tabular}}
\end{table*}

\end{document}